\documentclass[default]{sn-jnl}
\usepackage{multirow}%
\usepackage{amsmath,amssymb,amsfonts}%
\usepackage{amsthm}%
\usepackage{mathrsfs}%
\usepackage{tabularx}
\usepackage[linesnumbered, ruled, vlined]{algorithm2e}
\usepackage[title]{appendix}%
\usepackage{xcolor}%
\usepackage{textcomp}%
\usepackage{manyfoot}%
\usepackage{booktabs}%
\usepackage{listings}
\usepackage{graphicx}
\usepackage{color}
\usepackage{tabularx}
\usepackage{tabulary}
\usepackage{longtable}
\usepackage{booktabs} 
\usepackage{siunitx} 
\usepackage{caption}
\usepackage{booktabs}
\usepackage{siunitx}
\usepackage{makecell}
\usepackage{subcaption}
\usepackage{url}
\usepackage{setspace} 
\usepackage{lineno}
\usepackage{float}
\usepackage[markup=underlined]{changes}
\definechangesauthor[name={Editor}, color=blue]{Editor}

\begin{document}

\title{Improved Classification of Nitrogen Stress Severity in Plants Under Combined Stress Conditions Using Spatio-Temporal Deep Learning Framework}

\author[1,3]{\fnm{Aswini Kumar} \sur{Patra}}\email{aswinipatra@gmail.com / akp@nerist.ac.in}

\author[2]{\fnm{Anshu Rastogi}}\email{anshu.rastogi@up.poznan.pl}

\author*[3]{\fnm{Lingaraj} \sur{Sahoo}}\email{ls@iitg.ac.in}

\affil[1]{\orgdiv{Dept. of Computer Science \& Engineering}, \orgname{North Eastern Regional Institute of Science and Technology}, \country{Itanagar, India}}

\affil[2]{\orgdiv{Laboratory of Bioclimatology}, \orgname{Poznan University of Life Sciences}, \country{Piatkowska 94, Poznan  60-694, Poland}}

\affil[3]{\orgdiv{Dept. of Bio-Science and Bio-Engineering}, \orgname{Indian Institute of Technology Guwahati}, \country{India}}

\abstract{In the real-world scenario multiple stresses often appear together in the field such as drought and nutrient. As drought can inhibit nutrient uptake and increase water requirements for nutrient conversion, while imbalances in plant nutrients can also weaken cell membranes, increasing susceptibility to drought. This combined stress leads to synergistic negative effects on plant growth and yield, causing greater reductions in seed quality and biomass than either stress alone. Early detection of these stresses is therefore crucial for protecting plant health and implementing effective management strategies. This study proposes an improved deep learning framework to accurately classify nitrogen stress severity in a combined stress environment. Our model uses RGB, two stereo-infrared channels, and multi-spectral imagery to capture a wide range of physiological plant responses from canopy images. These images, provided as time-series data, document plant health across three levels of nitrogen availability (low, medium, and high) under varying water stress and weed pressures. The core of our approach is a spatio-temporal deep learning pipeline that merges a Convolutional Neural Network (CNN) for extracting spatial features from images with a Long Short-Term Memory (LSTM) network to capture temporal dependencies.  Two complementary evaluation strategies were used. First, a forward-chaining cross-validation protocol was adopted to assess temporal generalization to later acquisition dates while preserving chronological order. Second, a box-disjoint evaluation was performed to assess generalization to unseen cultivation boxes. Under forward-chaining cross-validation, the CNN--LSTM framework achieved a mean validation accuracy of 95.37\%, whereas the devised spatial-only framework achieved a mean validation accuracy of approximately 80.49\%, demonstrating the benefit of modelling temporal stress progression. Both approaches outperformed the box-disjoint evaluation, in which the model achieved a mean accuracy of 72.59\%, with performance varying across combined-stress configurations. These results indicate that spatio-temporal learning improves nitrogen-stress classification under combined-stress conditions, while also showing that generalization to unseen cultivation boxes remains challenging when nitrogen deficiency, water limitation, and weed pressure produce overlapping plant responses. Overall, the proposed framework provides a promising basis for temporally informed and deployment-oriented crop-stress monitoring in precision agriculture.}




\maketitle
\section{Introduction}
 Among all essential macro-nutrients, nitrogen (N) deficiency represents a major constraint on plant growth, development, and productivity \cite{hirel_improving_2011}. As a fundamental component of amino acids, proteins, nucleic acids, and chlorophyll \cite{singh_global_2022}, nitrogen plays a central role in multiple physiological and metabolic processes. Its deficiency disrupts these pathways, resulting in reduced leaf area, chlorosis, lower leaf count, and stunted plant height \cite{das2023long}. Beyond nutrient limitations, abiotic stressors such as drought and biotic pressures like weed competition frequently co-occur, compounding the negative effects on plant health. For example, water stress restricts nutrient mobility and uptake, thereby intensifying the impacts of nitrogen deficiency \cite{saud_effects_2017}. In natural environments, plants rarely face single stress factors in isolation. Rather, stress events often occur simultaneously or sequentially, interacting in synergistic or antagonistic ways \cite{blum_stress_2016, pandey2017impact}. These multi-stress combinations induce overlapping phenotypic symptoms, complicating efforts to diagnose the underlying causes \cite{webber2022framework}. 
Despite this, the majority of plant stress phenotyping studies have focused on single-stress scenarios, with relatively limited progress in understanding coexisting stresses \cite{mahalingam2014consideration, li2025enhancing, zagorvsvcak2025integration}. 
This gap demands the need for advanced tools capable of modeling the intricate, multi-dimensional dynamics of plant responses under combined stress conditions. 

The advent of imaging and sensor technologies has transformed plant research through the development of high-throughput phenotyping platforms \cite{Tamimi_2022}. These platforms, driven by computer vision and imaging-based phenomics, facilitate non-invasive and automated monitoring of key morphological and physiological traits \cite{yang_crop_2020}. Such techniques enable rapid, scalable, and effective assessment of plant health, significantly enhancing yield prediction and stress diagnosis capabilities \cite{humplik_automated_2015}. Machine learning (ML) \cite{gill_comprehensive_2022} and deep learning (DL) \cite{singh_deep_2018} have become indispensable tools in this domain, capable of capturing subtle, nonlinear patterns indicative of diverse stress conditions. Among DL models, Convolutional Neural Networks (CNNs) have demonstrated strong performance in extracting spatially significant features from RGB, hyperspectral, and thermal imagery \cite{jiang_convolutional_2020}. When integrated with temporal modeling frameworks like Long Short-Term Memory (LSTM) networks, these models can learn the progression of stress responses over time.

Recent research reflects a growing interest in nitrogen stress detection under both isolated and combined stress conditions. Clarke et al. \cite{clarke_spatial-temporal_2024} examines how spatial and temporal soil variability influences nitrogen use efficiency (NUE) in wheat using the Sirius crop simulation model and long-term field data. It finds that soil electrical conductivity (ECa) can guide site-specific nitrogen management, with lower water-holding soils requiring less nitrogen but posing higher leaching risks. Sarkar et al. \cite{sarkar_machine_2025} investigates how abiotic stressors—especially drought and temperature—affect nitrogen dynamics and crop productivity in dryland forage systems. Using field data and machine learning (ML) analysis, the study compares conventional tillage and no-till practices, along with the impact of green manures such as field peas. The results show that no-till systems with green manuring significantly improve nitrogen use efficiency (NUE) and reduce the negative effects of drought on plant growth. Combining SPAD data from multiple leaf positions significantly improves the estimation of the Nitrogen Nutrition Index (NNI), as demonstrated in another study by Wang et al. \cite{wang_enhancing_2024}, where machine learning models like Random Forest and XGBoost outperformed linear regression in predicting NNI. A spatio-temporal spectral framework combining RGB, infrared, hyperspectral data and derived plant traits like canopy cover, height, biomass, and vegetation indices to detect drought, nitrogen, and weed stress in sugar beet. Machine learning models, especially SVM, showed high accuracy with multi-modal features outperforming single ones \cite{khanna_spatio_2019}. 
A reinforcement learning (RL) environment was developed by Kallenberg et al. \cite{kallenberg_nitrogen_2023} where agents learn crop management policies through crop growth models. In a nitrogen management case study for winter wheat, the RL agent successfully detected crop nitrogen requirements by analyzing growth states and guided optimal fertilizer application. Ghazal et al. evaluates machine learning models for nitrogen stress detection in maize using RGB images under field conditions. Among tested models, EfficientNetB0 achieved the highest accuracy, outperforming vision transformers and other CNNs\cite{ghazal_comparative_2024}. A study developed machine learning and deep learning models for image-based nitrogen diagnosis in muskmelon using canopy leaf images and environmental data. Among all models they devised, the hybrid DCNN–LSTM achieved the highest accuracy by combining spatial features and temporal light–temperature inputs \cite{chang_using_2021}. A hybrid deep learning model is proposed by Liao et al. that integrates CNN with an attention mechanism and LSTM to diagnose nitrogen (N) and potassium (K) nutrient levels in rice at the early panicle initiation stage \cite{liao_hybrid_2024}. The study by Hui et al. estimated sugarcane nitrogen levels using digital images and regression-based machine learning models, including Random Forest (RF), Backpropagation Neural Network (BPNN), and a stacking fusion approach. Results showed that the fusion model with PCA-based color–texture features outperformed both RF and BPNN \cite{you_sugarcane_2023}. Electrophysiological signals proved to be a successful modality for detecting nitrogen deficiency stress in tomato plants grown under greenhouse conditions, with deep learning—particularly an encoder-based architecture—outperforming models such as XGBoost \cite{gonzalez_i_jucla_detecting_2023}.
Chaparro et al. estimated foliar nitrogen content in pineapple by integrating multi-spectral UAV imagery, IoT-based environmental sensors, and SPAD chlorophyll values with machine learning. Of the nine models tested, XGBoost and multi-layer perceptron (MLP) achieved the highest accuracies, while multi-sensor data fusion consistently outperformed image-only approaches \cite{chaparro_machine_2024}. Hyperspectral remote sensing is combined with stepwise multiple linear regression to detect nitrogen and water stress in maize in a study by Naik et al. Nitrogen stress was most effectively identified at 540, 780, and 860 nm, with leaf nitrogen content accounting for up to 66\% of yield variation at the tasseling stage \cite{naik_identification_2020}. Trung-Tin Tran et al. \cite{tran_comparative_2019} employ  two distinct models, namely Inception-ResNet v2 and an Autoencoder based on convolutional neural networks, to classify and predict nutrient deficiency symptoms, specifically related to calcium, potassium, and nitrogen. Azimi et al. developed a 23-layer CNN to classify nitrogen deficiency stress in sorghum using shoot images. It outperformed classical ML methods and performing comparably to deeper models like ResNet18 and NasNet Large with far fewer parameters \cite{azimi_deep_2021}. In summary, the literature uses single and multi-modal datasets that include imaging and spectral, physiological, biochemical, environmental, electrophysiological, and visual trait data—often processed with machine learning and deep learning techniques to improve nitrogen stress detection and management.

Despite these advances, most studies examine stresses in isolation, overlooking the complex interactions that occur when nutrient stress coincides with other environmental pressures. This gap limits the applicability of current models in real-world conditions, where stresses such as nitrogen deficiency, drought, and weed pressure often co-occur. To address this challenge, we propose a spatio-temporal deep learning framework that leverages pre-trained CNNs in combination with LSTMs to capture both spatial features and temporal growth dynamics, enabling accurate classification of nitrogen stress severity under combined drought and weed pressure. The novelty lies in the combined-stress application, multi-modal feature processing, lightweight architecture, and temporally ordered evaluation.

The study was guided by the following scientific questions:
\begin{itemize}[leftmargin=2em]
    \item Can nitrogen stress severity be classified from canopy images when nitrogen deficiency occurs together with drought and weed pressure?

    \item Does modelling temporal progression across acquisition dates improve nitrogen-stress classification compared with a spatial-only CNN model?

    \item How does the model perform across successive growth-stage intervals and when evaluated on unseen cultivation boxes?
\end{itemize}

Based on these questions, we tested the following hypotheses:
\begin{itemize}[leftmargin=2em]
    \item \textbf{H1:} A spatio-temporal CNN--LSTM framework will outperform a spatial-only CNN baseline because temporal image sequences contain stress-progression information that is not available in single-date images.

    \item \textbf{H2:} Model performance will vary across evaluation settings, with stronger performance expected under temporally ordered forward-chaining validation and more challenging performance under box-disjoint validation, where the model must generalize to previously unseen cultivation boxes.
\end{itemize}

The key contributions of this study are as follows:
\begin{itemize}
    \item We developed a lightweight CNN--LSTM framework that integrates 
    MobileNetV2-based spatial feature extraction with LSTM-based temporal 
    modeling to classify low, medium, and high nitrogen stress severity under 
    combined drought and weed-pressure conditions.

    \item We evaluated three lightweight CNN backbones, namely MobileNetV2, 
    EfficientNetB0, and NASNetMobile, for spatial feature extraction. Among 
    these models, MobileNetV2 provided the best performance and was therefore 
    selected as the feature extractor in the proposed CNN--LSTM framework.

    \item We implemented two complementary evaluation strategies: a 
    forward-chaining protocol to assess temporal generalization across 
    successive acquisition dates and growth-stage intervals, and a box-disjoint 
    protocol to assess generalization to previously unseen cultivation boxes. 
    These strategies provide a more 
    realistic assessment of model performance.

    \item We compared the proposed spatio-temporal framework with a spatial-only 
    MobileNetV2 baseline and conventional machine-learning approaches, thereby 
    demonstrating the added value of temporal modeling for nitrogen-stress 
    classification under combined stress conditions.
\end{itemize}

\section{Materials and Methods}
In this section, we first introduce the experimental dataset used in the study. We then describe the proposed spatio-temporal framework, followed by the spatial-only network architecture. Finally, we outline the performance metrics employed for model evaluation.


\subsection{Data Description}

\begin{table*}[h]
    \centering
    \caption{Combined Stress Treatment}
    \label{stress}
    \begin{tabular}{|c|c|c|c|}
        \hline
        \textbf{Nitrogen Input} & \textbf{Water Input} & \textbf{Weed Pressure} & \textbf{Box Numbers} \\
        \hline
        Low & Sufficient & None & 22,23,24 \\
        \hline
        Medium & Sufficient & None & 4,5,6 \\
        \hline
        High & Sufficient & Medium & 7,8,9 \\
        \hline
        High & Sufficient & High & 10,11,12 \\
        \hline
        Medium & Low & None & 13,14,15 \\
        \hline
        High & Sufficient & High & 16 \\
        \hline
        High & Sufficient & High & 17 \\
        \hline
        High & Sufficient & High & 18 \\
        \hline
        Low & Sufficient & Medium & 25,26,27 \\
        \hline
        Medium & Low & High & 19,20,21 \\
        \hline
        Low & Low & None & 28,29,30 \\
        \hline
    \end{tabular}
\end{table*}

The dataset utilized in this study is derived from the work by Khanna et al.\citep{khanna_spatio_2019, data_set_18}, who established a comprehensive plant phenotyping framework to investigate the physiological effects of combined abiotic (drought) and biotic (weed competition) stresses alongside nitrogen deficiency in sugar beet (Beta vulgaris L.). Their experimental design closely mimicked field-realistic stress scenarios, enabling systematic evaluation of plant responses under factorial combinations of low, medium, and high nitrogen supply, with varying water availability and weed presence. This design aimed to suggest the complex interactions between multiple, simultaneously occurring stressors, which often induce overlapping phenotypic responses such as reductions in leaf area, biomass, and visible symptoms like chlorosis.To classify nitrogen stress levels in sugar beet plants, we utilized canopy images from multiple modalities, namely RGB, infrared, and multi-spectral. 
The images were collected using an Intel® RealSense™ ZR300 camera—providing RGB and dual infrared (stereo IR) channels and a Ximea MQ022HG-IM-SM5X5 camera capturing multi-spectral images.

\begin{figure*}[hbt!]
    \centering
    
    \begin{subfigure}{0.45\textwidth}
    \centering
        \includegraphics[width=0.9\textwidth]{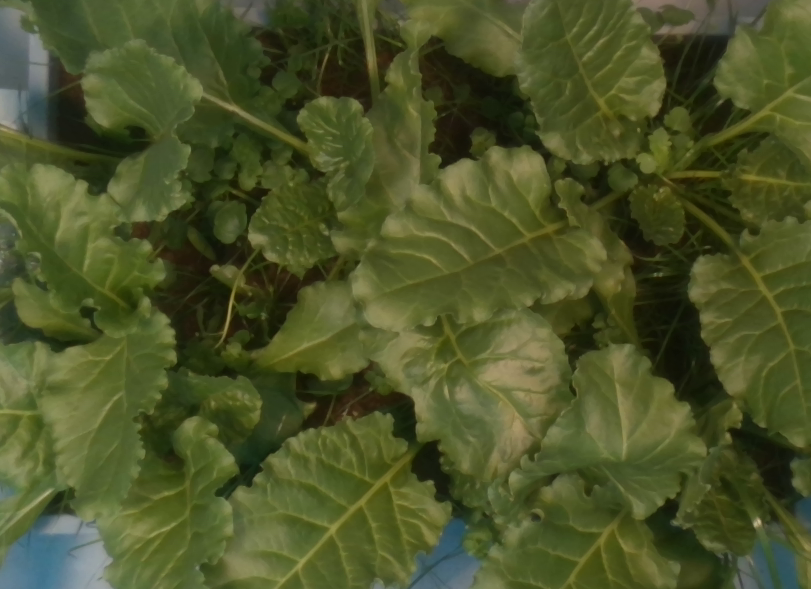}
        \caption{Low nitrogen (sufficient water, medium weed), RGB}
        \label{fig:low_nitrogen}
    \end{subfigure}
    \hspace{0.5cm}
    \begin{subfigure}{0.45\textwidth}
    \centering
        \includegraphics[width=0.9\textwidth]{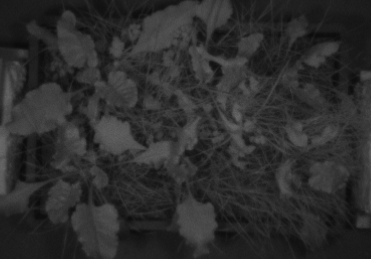}
        \caption{Medium nitrogen (low water, high weed), Infra Red 1 }
        \label{fig:medium_nitrogen1}
    \end{subfigure}

    \vspace{0.5cm}
    
    \begin{subfigure}{0.45\textwidth}
    \centering
        \includegraphics[width=0.9\textwidth]{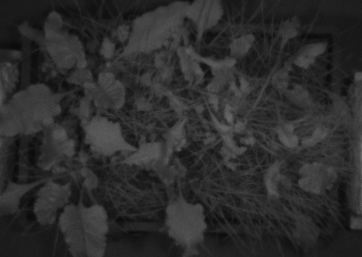}
        \caption{Medium nitrogen (low water, high weed), Infra Red 2}
        \label{fig:medium_nitrogen2}
    \end{subfigure}
    \hspace{0.5cm}
    \begin{subfigure}{0.45\textwidth}
    \centering
        \includegraphics[width=0.9\textwidth]{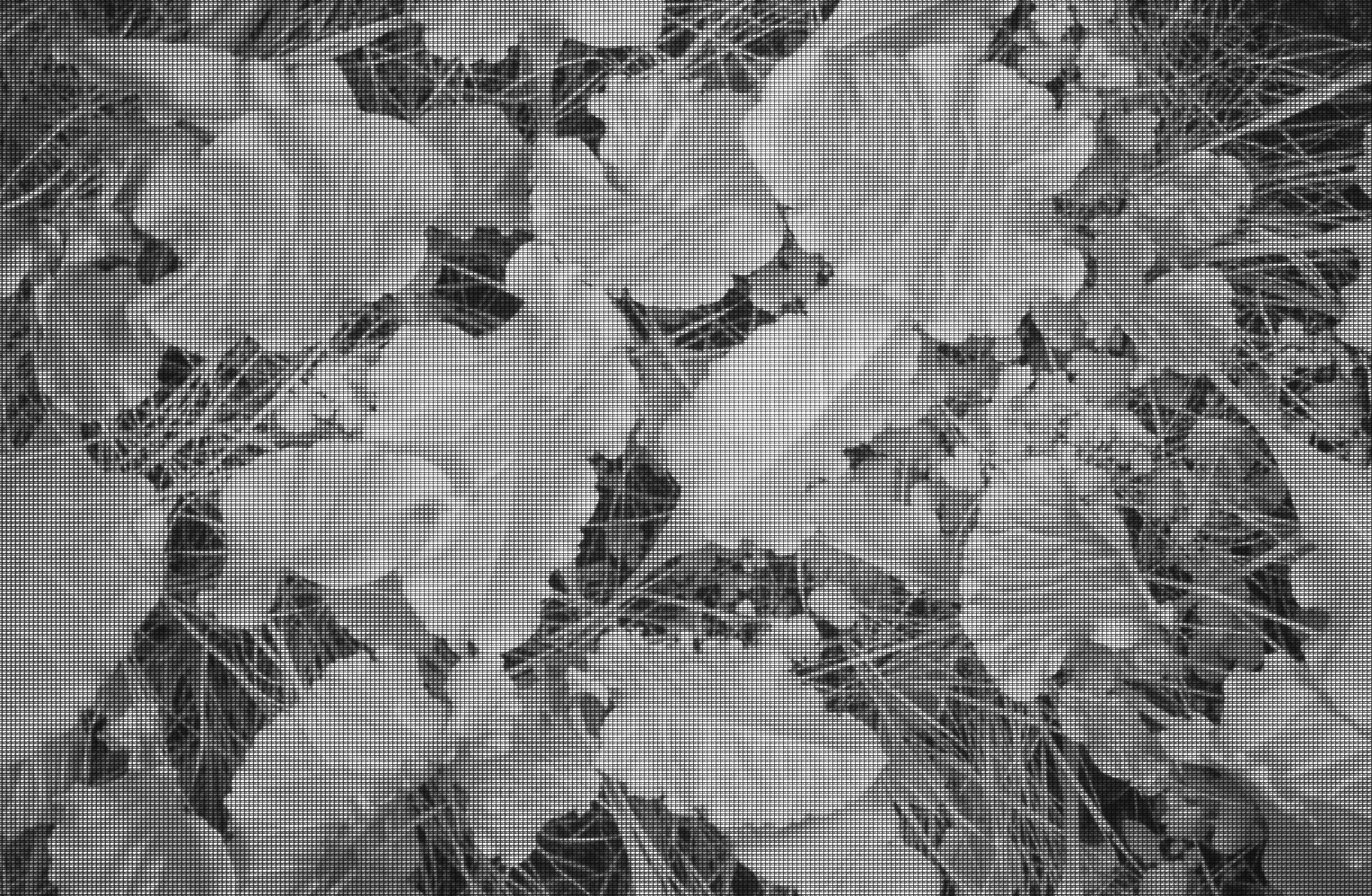}
        \caption{High nitrogen (sufficient water, high Weed), Multi-Spectral}
        \label{fig:high_nitrogen}
    \end{subfigure}

    \caption{Nitrogen deficiency levels(with varying levels of water and weed) on a specific day captured by RGB, Infra Red and Multi-spectral sensor.}
    \label{fig:nitrogen_deficiency}
\end{figure*}

Each stress factor was applied at different severity levels. Nitrogen availability was assessed using three levels—low, medium, and high—representing deficient, sufficient, and surplus nitrogen supply, equivalent to 20, 40, and 80 kg/ha, respectively. Weed pressure was categorized as no weeds, medium pressure (chickweeds), or high pressure (chickweeds and grasses). Water supply was manually regulated at two levels—sufficient and low. As a result, plants experienced varying combinations of these three stressors at any given time. The experiment was intentionally structured to emulate real-world conditions by applying nitrogen deficiency, drought, and weed competition both individually and in combination. Nitrogen deficiency levels with varying water and weed pressure captured by RGB, stereo infrared, and multi-spectral sensors on a specific day are illustrated in Fig. \ref{fig:nitrogen_deficiency}. The detailed stress treatments are summarized in Table \ref{stress}, which lists the combinations of nitrogen input, water input, and weed pressure, along with the corresponding cultivation box numbers. The treatment matrix (i.e., Table \ref{stress}) assigns 27 cultivation boxes to these combinations. The dataset includes images from 16 measurement dates throughout the growth period, featuring 27 boxes (9 per nitrogen level). Images from 14 dates were retained for analysis, excluding the first two dates due to early-stage germination where stress symptoms were minimal. The 14 acquisition dates used in our analysis span the period from 1 February 2018 to 29 March 2018, covering approximately 57 days of the sugar beet growth cycle. For each nitrogen level category, nine boxes were imaged across four modalities, yielding 504 images per category (14 dates × 9 boxes × 4 modalities). Across all three nitrogen levels, a total of 1,512 images, collected in a controlled environment, were used in this study. All images were cropped to remove irrelevant background content.

\subsection{Proposed Framework}

\subsubsection{Spatio-Temporal Framework}

\begin{figure*}[!h]
        \centering
		\includegraphics[scale=0.72]{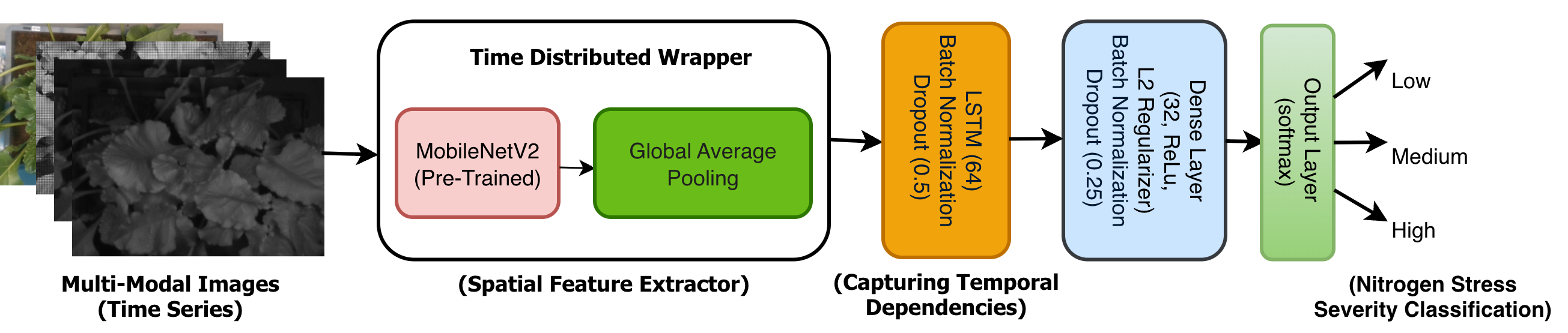}
		\caption{Spatio-Temporal Deep Learning Framework for Nitrogen Stress Severity Classification} \label{frame_cnn_lstm}
\end{figure*}

In this study, we present a deep learning framework that integrates spatial and temporal features to classify images into three nitrogen severity levels: low, medium, and high. The model integrates a CNN for feature extraction and a LSTM network for temporal sequence modeling. The overall architecture of the
proposed framework is depicted in Fig. \ref{frame_cnn_lstm}. The architecture of the proposed CNN-LSTM hybrid consists of the following components:

\begin{enumerate}

    \item \textbf{Feature Extractor:} A pre-trained \texttt{MobileNetV2} (with \texttt{imagenet} weights) served as the base CNN, where the classification head was removed. The network's output was passed through a Global Average Pooling layer to obtain a fixed-size feature vector per image. This CNN was wrapped within a \texttt{TimeDistributed} layer to process each frame of the sequence independently while sharing weights. The rationale for selecting \textit{MobileNetV2} \cite{sandler2018mobilenetv2} as the feature extractor is its lightweight architecture, with only 3.4 million parameters. Its accuracy as a baseline classifier is higher compared with other lightweight models, such as \textit{NASNetMobile} \cite{zoph2018learning} and \textit{EfficientNetB0} \cite{tan2019efficientnet}, each having approximately 5.3 million parameters. A lightweight model reduces computational complexity, requires less memory,  and is suitable for deployment on resource-constrained devices.
    
    \item \textbf{Temporal Encoder:} The sequence of image features was then fed into an LSTM layer with 64 hidden units to learn temporal patterns across the sequence.

    LSTM networks are an extension of recurrent neural networks (RNNs) designed to address the vanishing gradient problem and effectively capture long-term dependencies in sequential data \cite{hochreiter1997long}. In LSTM models, a memory cell with gating mechanisms enables the network to retain and utilize information over extended sequences, allowing for the reading, writing, and deletion of information from its memory. These gating mechanisms, comprised of forget, input, and output gates, play crucial roles in managing the flow of information within the LSTM unit \cite{gers2000learning}. An LSTM unit consists of three main components:

\begin{enumerate}
    \item \textbf{Forget Gate ($f_t$)}: Evaluates the relevance of existing information stored in the memory cell. It decides which information to retain and which to discard based on the input at the current time step ($x_t$) and the previous hidden state ($h_{t-1}$). Mathematically, the output of the forget gate ($f_t$) is computed using a sigmoid activation function:
    
  \[
f_{t}=\sigma\left(W_{f_{h}}h_{t-1} + W_{f_{x}}x_{t} + b_{f}\right) \tag{1}
\]

where $W_{f_{h}}$ and $W_{f_{x}}$ are weight matrices, and $b_{f}$ is the bias. The symbol \( \sigma \) represents the sigmoid activation function, which is defined as:

\[
\sigma(x) = \frac{1}{1 + e^{-x}}
\]

where \( e \) is the base of the natural logarithm, and \( x \) is the input to the sigmoid function.

\item \textbf{Input Gate ($i_t$) and Candidate Cell State ($\tilde{c}_t$)}: Determines how much new information should be added to the memory cell. It consists of a sigmoid layer that controls the update and a "tanh" layer that generates a vector of new candidate values. The input gate output ($i_t$) and the candidate cell state ($\tilde{c}_t$) are computed as follows:

\[
i_t=\sigma(W_{ih}h_{t-1} + W_{ix}x_{t} + b_{i}) \tag{2}
\]

\[
\tilde{c}_t=\tanh(W_{ch}h_{t-1} + W_{cx}x_{t} + b_{c}) \tag{3}
\]

The candidate cell state represents the new information to be added to the memory cell.

\item \textbf{Memory Update and Output Gate}: Updates the memory cell content based on the forget gate output ($f_t$), input gate output ($i_t$), and candidate cell state ($\tilde{c}_t$). The updated cell state ($c_t$) is calculated as:

\[
c_t=f_t \odot c_{t-1} + i_t \odot \tilde{c}_t \tag{4}
\]

where \( \odot \) denotes element-wise multiplication.

Finally, the output gate controls which parts of the cell state contribute to the output. The output gate output ($o_t$) and the final hidden state ($h_t$) are computed as:

\[
o_{t}=\sigma\left(W_{o h}h_{t-1} + W_{o_{x}}x_{t} + b_{o}\right) \tag{5}
\]

\[
h_{t}=o_{t} \odot \tanh \left(c_{t}\right) \tag{6}
\]

The output gate output ($o_t$) determines the relevance of the current cell state, and the final hidden state ($h_t$) represents the LSTM's output at the current time step.
\end{enumerate}

In summary, LSTM models utilize gated memory cells to effectively capture and retain long-term dependencies in sequential data, addressing the limitations of traditional RNNs. The forget, input, and output gates enable the LSTM to selectively process and utilize information, making it a powerful tool for tasks involving sequential data analysis and prediction.

    \item \textbf{Fully Connected Layers:} The output of the LSTM layer (64 units) was followed by Batch Normalization and Dropout (0.5), then passed through a Dense layer with 32 units and ReLU activation with L2 regularization, followed by another Batch Normalization and Dropout (0.25), and finally an output layer with softmax activation for multiclass classification.
\end{enumerate}

\begin{algorithm*}
\SetAlgoLined
\DontPrintSemicolon

\KwData{%
    Image dataset from \texttt{DATA\_DIR} organised into per-class subdirectories;
    sequence length $T = 5$;
    samples per window $K = 36$;
    epochs $= 60$;
    batch size $= 16$%
}
\KwResult{%
    Per-fold learning curves, classification reports, confusion matrices,
    and aggregated cross-validation summary%
}

\BlankLine
\tcp{\textbf{Phase 1: Dataset construction}}
Scan \texttt{DATA\_DIR}; parse 8-character date prefix from each filename\;
Build DataFrame \texttt{df} $\leftarrow$ $\langle$\texttt{filepath, class, date, group\_id}$\rangle$\;

\BlankLine
\tcp{\textbf{Phase 2: Temporal sequence construction}}
\For{\emph{each} $(\texttt{class},\ \texttt{group})$ $\in$ \texttt{df.groupby("class")}}{
    Slide window of length $T$ over sorted unique dates;
    sample $K$ sequences per window;
    record \texttt{last\_date} for each sequence\;
}
Sort all sequences chronologically by \texttt{last\_date};
encode class labels\;

\BlankLine
\tcp{\textbf{Phase 3: Forward-chaining cross-validation}}
Set $\texttt{min\_anchor\_idx} \leftarrow T + 4 - 1 = 8$;
derive six eligible \texttt{val\_anchor} dates\;

\For{\emph{each} \texttt{val\_anchor} $\in$ \texttt{foldable\_dates}}{

    \tcp{Temporal partition}
    $\texttt{train} \leftarrow \{\,i : \texttt{last\_date}_i < \texttt{val\_anchor}\,\}$;\quad
    $\texttt{val} \leftarrow \{\,i : \texttt{last\_date}_i = \texttt{val\_anchor}\,\}$\;
    \lIf{$|\texttt{train}| < 100$ \textbf{or} $|\texttt{val}| = 0$}{\textbf{skip} fold}

    \BlankLine
    \tcp{Image loading and augmentation}
    $\mathbf{X}_\text{train} \leftarrow \texttt{load\_seq\_batch}(\texttt{train})$
    \tcp*{shape: $(N_\text{train},\,T,\,224,\,224,\,3)$}
    $\mathbf{X}_\text{val} \leftarrow \texttt{load\_seq\_batch}(\texttt{val})$
    \tcp*{raw only; never augmented}
    One-hot encode $\mathbf{y}_\text{train}$ and $\mathbf{y}_\text{val}$ to $C$ classes\;
    \For{\emph{each} sequence $i$ in $\mathbf{X}_\text{train}$}{
        Sample one random spatial transform $\boldsymbol{\tau}^{(i)}$
        (rotation, flip, shift, shear)\;
        Apply $\boldsymbol{\tau}^{(i)}$ identically to all $T$ frames of sequence $i$
        \tcp*{preserves temporal coherence}
    }
    $\mathbf{X}_\text{train}^{\text{final}} \leftarrow
        \texttt{concat}\bigl[\mathbf{X}_\text{train},\ \mathbf{X}_\text{train}^{\text{aug}}\bigr]$;\quad
    $\mathbf{y}_\text{train}^{\text{final}} \leftarrow
        \texttt{concat}\bigl[\mathbf{y}_\text{train},\ \mathbf{y}_\text{train}\bigr]$
    \tcp*{training set doubled}

    \BlankLine
    \tcp{Model construction (fresh each fold)}
    Instantiate \texttt{MobileNetV2} (ImageNet weights); freeze all convolutional layers\;
    Attach \texttt{GlobalAveragePooling2D} $\Rightarrow$ spatial extractor
    $f_\theta : \mathbb{R}^{224\times224\times3} \to \mathbb{R}^{d}$\;
    Wrap $f_\theta$ in \texttt{TimeDistributed} to process all $T$ frames independently\;
    Pass frame-level features through $\texttt{LSTM}(64,\ \texttt{return\_sequences=False})$\;
    Stack \texttt{BN} $\to$ \texttt{Dropout}$(0.5)$ $\to$
          \texttt{Dense}$(32,\,\texttt{relu},\,\ell_2)$ $\to$
          \texttt{BN} $\to$ \texttt{Dropout}$(0.25)$ $\to$
          \texttt{Dense}$(C,\,\texttt{softmax})$\;
    Compile with \texttt{Adam} (exponential decay, $\eta_0{=}10^{-4}$)
    and categorical cross-entropy\;

    \BlankLine
    \tcp{Training}
    Configure \texttt{EarlyStopping} (patience $= 10$, monitor \texttt{val\_loss})
    and \texttt{ModelCheckpoint} (best \texttt{val\_loss})\;
    Fit model on $(\mathbf{X}_\text{train}^{\text{final}},\,\mathbf{y}_\text{train}^{\text{final}})$;
    validate on $(\mathbf{X}_\text{val},\,\mathbf{y}_\text{val})$;
    save and plot loss and accuracy curves\;

    \BlankLine
    \tcp{Evaluation}
    Predict $\hat{\mathbf{y}}_\text{val}$ from the model on $\mathbf{X}_\text{val}$\;
    Compute fold accuracy, per-class F1, and confusion matrix\;
    Plot confusion matrix\;
}

\BlankLine

\caption{CNN--LSTM under Forward-Chaining Cross-Validation for Nitrogen Stress Classification}
\label{algo:cnn_lstm}
\end{algorithm*}

Algorithm \ref{algo:cnn_lstm} provides an overview of the complete workflow for constructing temporal image sequences, training the proposed model, and evaluating its performance using forward-chaining cross-validation. First, the canopy images are organized according to nitrogen class, acquisition date, and experimental group. A sliding window spanning five consecutive dates is then used to generate temporally ordered sequences, with 36 sequences sampled per nitrogen class for each window. These sequences are arranged chronologically according to their terminal date. In each fold, all sequences ending before the selected validation anchor are included in the training set, whereas sequences ending on the anchor date are retained for validation. This procedure preserves the temporal structure of the dataset and prevents information from future observations from entering the training process. Data augmentation is applied only to the training sequences, and the same randomly selected spatial transformation is applied to all five frames within a sequence to maintain temporal consistency. A new model is initialized for each fold using a frozen \textit{ImageNet}-pretrained \textit{MobileNetV2}backbone followed by global average pooling. The feature extractor is wrapped within a \texttt{TimeDistributed} layer, allowing the same \textit{MobileNetV2}-based feature extraction process to be applied independently to each frame while sharing weights across all time steps. The resulting sequence of frame-level feature vectors is then passed to a 64-unit LSTM, and its final hidden representation is processed through batch normalization, dropout, and a 32-unit dense layer with L2 regularization before classification into the three nitrogen-severity classes using a softmax output layer. The model is trained using the \textit{Adam} optimizer with an exponentially decaying learning rate, together with early stopping and model checkpointing based on the minimum validation loss. Finally, performance is assessed in each fold using classification accuracy, class-specific F1-scores, confusion matrices, and learning curves.

\subsubsection{Evaluation Procedures}

Two CNN–LSTM experiments were conducted to examine complementary aspects of temporal nitrogen-stress classification. Both experiments used \textit{MobileNetV2} to extract spatial features from five-frame image sequences and an LSTM to model information across acquisition dates. However, they differed in sequence construction and validation strategy and therefore addressed different generalization questions.

In the class-level forward-chaining approach, five-frame sequences were generated by randomly selecting one image from each of five consecutive dates within the same nitrogen class. Box and modality identities were not preserved. Sequences were split according to their terminal date, with earlier sequences used for training and later-anchor sequences used for validation as shown in Algorithm \ref{algo:cnn_lstm}. 

In the box-disjoint approach, sequences were constructed within individual cultivation boxes across five consecutive dates. Entire boxes were then assigned exclusively to training, and test set, ensuring that no box or image file was shared across partitions, as demonstrated in Algorithm \ref{alg:lobo}. Because modality was sampled independently at each date, modality identity was not necessarily constant within a sequence. This experiment evaluated generalization to previously unseen cultivation boxes.

The two analyses therefore address complementary questions: the first measures temporal generalization to later dates, whereas the second measures identity generalization to unseen boxes. Their results were interpreted separately because they use different sequence-construction and validation strategies.

\textbf{Forward-Chaining Cross-Validation}

The class-level forward-chaining approach was designed to examine whether nitrogen-stress signatures remain detectable across successive acquisition dates (i.e growth stages). To construct temporally ordered input
sequences, a sliding window of length $T = 5$ consecutive dates was applied
independently to each stress class. The window length was determined through
preliminary experiments with different sequence lengths, and ($T = 5$) was
found to provide the best model performance.
Within each window, $K = 36$ sequences were generated per class by random
sampling of one image per date, yielding a total sequence pool sorted
chronologically by the terminal date of each window (hereafter the
\textit{last\_date}). Under this scheme, each fold is defined by a single
\textit{validation anchor} date, selected from the set of eligible dates whose
index satisfies the constraint:

\[
\text{min\_anchor\_idx}
= T + \text{MIN\_TRAIN\_DATES} - 1
= 5 + 4 - 1
= 8
\tag{7} \label{eq:min-anchor}
\]

\begin{figure*}
    \centering
    \includegraphics[width=1\textwidth, keepaspectratio]{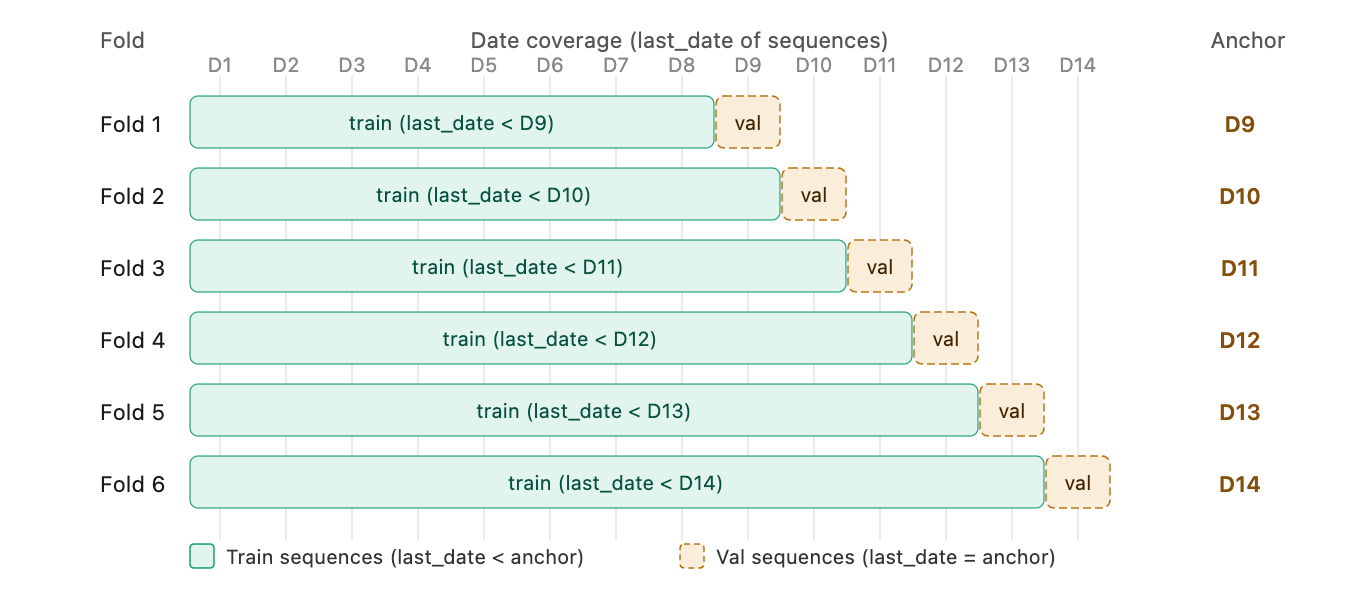}
    \caption{Forward-chaining Cross-Validation}
    \label{Forward_chain}
\end{figure*}

This constraint guarantees that at least four training windows
($4 \times 36 \times 3 = 432$ raw sequences prior to augmentation) precede the
earliest possible validation anchor. Given 14 unique collection dates
$(D_1, D_2, \ldots, D_{14})$, this produces six eligible validation anchors and
correspondingly six folds, as illustrated in Fig.~\ref{Forward_chain}. Because
each window is identified by its terminal date, window $W_k$ ends on date
$D_{k+4}$; the first eligible anchor $D_9$ therefore corresponds to window
$W_5$, and windows and their anchor dates may be referred to interchangeably.

For each fold, the partition rule is exact and non-overlapping: all sequences
whose \textit{last\_date} is strictly less than the validation anchor are
assigned to the training set, while sequences whose \textit{last\_date} equals
the validation anchor constitute the validation set. Because the split is drawn
on \textit{last\_date} alone---the date of a sequence's fifth and most recent
frame---no individual image can appear in both a training and a validation
sequence: every training sequence terminates strictly before the anchor, so no
training frame is dated on or after the anchor. The anchor date, which carries
the classification target, is therefore never seen during training. The earlier
frames of a validation sequence share dates with training sequences by design;
this common historical context is intrinsic to forward-chaining validation and
is distinct from leakage, in which the target-bearing observation itself is
duplicated across the split.
The validation set size is fixed at 108 sequences per fold
($1~\text{window} \times 36~\text{samples} \times 3~\text{classes}$), while
the training set grows with each successive fold as the anchor advances
(Fig.~\ref{Forward_chain}).

\begin{figure*}
    \centering
    \includegraphics[width=1\textwidth, keepaspectratio]{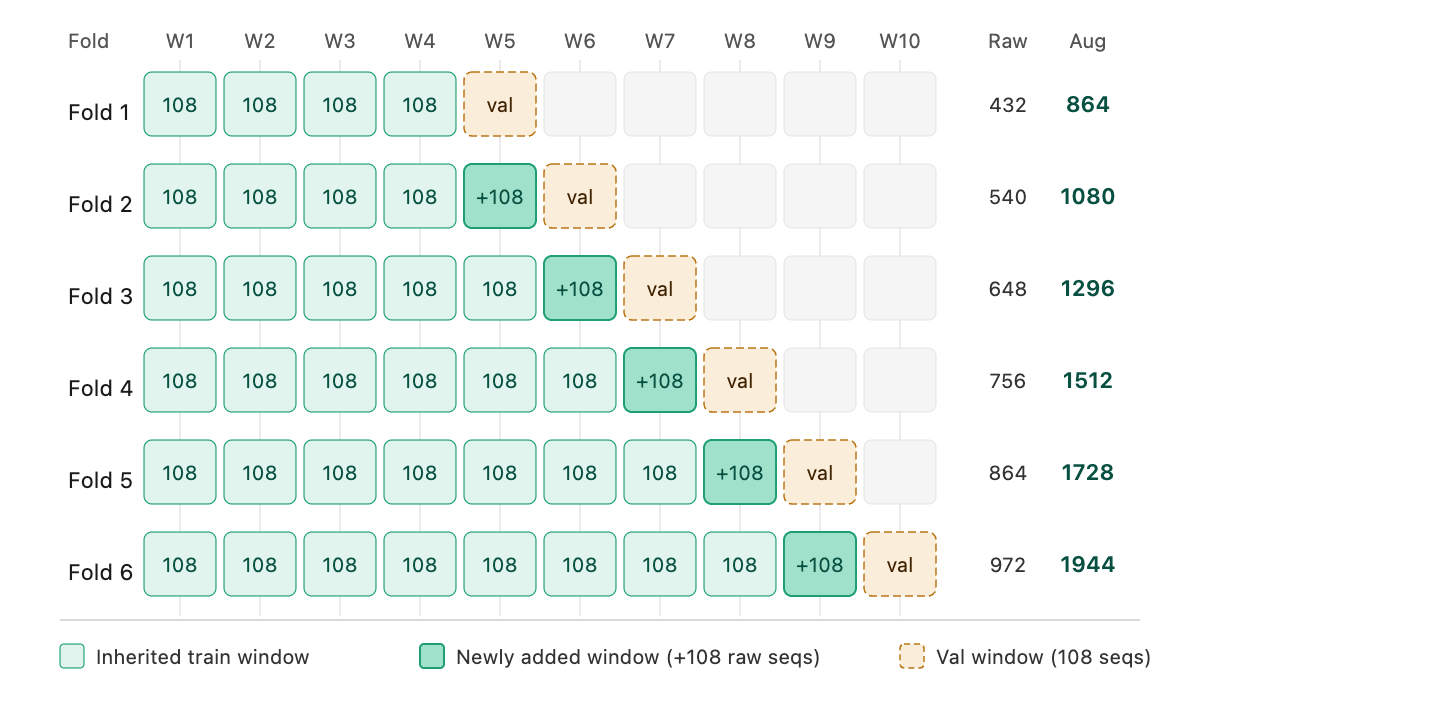}
    \caption{Training Window Accumulation Across Folds}
    \label{training_window}
\end{figure*}

A defining property of the forward-chaining cross-validation strategy is that
the training set grows strictly and monotonically with each successive fold,
as illustrated in Fig.~\ref{training_window}. This behavior arises from three
structural properties of the partitioning scheme.

A sliding window of length $T = 5$ applied to 14 unique collection dates
yields $n_{\text{windows}} = 14 - T + 1 = 10$ windows in total ($W_1$ through
$W_{10}$, ending on dates $D_5$ through $D_{14}$ respectively). The minimum
anchor constraint (Eq.~\ref{eq:min-anchor}) reserves the first four windows
($W_1$--$W_4$, ending $D_5$--$D_8$) exclusively for training at the earliest
fold, leaving six windows ($W_5$--$W_{10}$, ending $D_9$--$D_{14}$) as eligible
validation anchors---one per fold. For each fold $f \in \{1, 2, \ldots, 6\}$,
the training set contains $W_0 + f - 1$ windows, where $W_0 = 4$, and the
validation set contains exactly one window (the $f$-th eligible anchor). At
Fold~6, windows $W_1$ through $W_9$ constitute the training set and window
$W_{10}$ serves as the validation set.

First, the split condition $\textit{last\_date} < \text{anchor}$ is a strict
inequality. As the validation anchor advances by one date per fold, every
sequence that constituted the validation set in fold $f$ automatically
satisfies the training condition in fold $f+1$, since its \textit{last\_date}
is now strictly less than the new anchor. Consequently, no previously used
sequence is ever discarded from the training pool.

Second, each advancing anchor unlocks exactly one additional sliding
window---the window whose terminal date equals the previous validation
anchor---adding $K \times C = 36 \times 3 = 108$ raw sequences to the training
set at each step.

Third, the training sequences are augmented by a factor of two prior to model
fitting, through the application of spatially consistent random transforms
across all frames in each sequence. This doubles the effective contribution of
each newly admitted window, such that the augmented training set grows by
exactly 216 sequences per fold.

The combined effect is a training set of size

\[
N_{\text{train}}(f)
= 2 \times \bigl(W_0 + f - 1\bigr) \times K \times C
\tag{8}
\]
where $W_0 = 4$, $K = 36$, $C = 3$, and $f \in \{1, 2, \ldots, 6\}$.

This yields augmented training set sizes of 864, 1080, 1296, 1512, 1728, and
1944 sequences across Folds~1 through~6 respectively, while the validation set
remains fixed at 108 sequences in every fold. This monotonic growth is a
deliberate design choice: later folds present the model with a larger and more
diverse temporal history, enabling assessment of whether performance improves
as training data accumulates---a practically important question for operational
deployment of the model on future growing seasons where only recent historical
data would be available.


%


\newpage

\textbf{Box-disjoint Approach}

The box-disjoint approach evaluates whether nitrogen-stress patterns learned from boxes exposed to different water and weed stress combinations can generalize to unseen boxes. Treatment-wise analysis can further reveal which nitrogen-severity levels are predicted most reliably under specific combined-stress conditions. 

As described in Algorithm \ref{alg:lobo}, the image were first grouped according to nitrogen class, cultivation box, and acquisition date. Images were collected on 14 acquisition dates for each cultivation box. These dates were arranged chronologically, and temporal windows containing 5 consecutive acquisition dates were generated. Thus, a maximum of 10 (14-5+1=10) temporal windows could be obtained from each cultivation box. For every temporal window, 4 (i.e. $k$ samples per box) temporal sequences were constructed by randomly selecting one available image from each of the five acquisition dates, irrespective of imaging modality. Therefore, each cultivation box could contribute a maximum of 40 ($10 \times 4=40$) temporal sequences. Each sequence represented the temporal development of the same cultivation box across five consecutive acquisition dates. Finally, the training dataset contains 960 sequences (i.e. $24 \times 40$) and test set contains 120 sequences (i.e. $3 \times 40$).

A stratified leave-one-box-out cross-validation procedure was subsequently applied. In each fold, one cultivation box from each nitrogen class was selected for testing, while all the remaining boxes were used for training. Since each nitrogen class contained nine cultivation boxes, the procedure resulted in nine evaluation folds. All temporal sequences and image files originating from the selected test boxes were assigned exclusively to the test set, whereas sequences from the remaining boxes were assigned to the training set. This ensured complete separation at both the cultivation-box and image-file levels and prevented information leakage caused by the same box or image appearing in both the training and test sets. The inclusion of one test box from each nitrogen class also made the test set approximately class-balanced in every fold. For each fold, a newly initialized CNN--LSTM model was trained independently so that information learned in one fold was not transferred to another. The final 10 layers of the ImageNet-pretrained MobileNetV2 backbone were retained as trainable, while all preceding layers were frozen. 

\begin{algorithm}[t]
\caption{Leave-One-Box-Out (per-class) Cross-Validation for Box-Disjoint Evaluation}
\label{alg:lobo}
\DontPrintSemicolon
\KwData{Image set from \texttt{DATA\_DIR}, organised as \texttt{DATA\_DIR/<class>/<date>\_<box>\_<modality>.jpg};
sequence length $T=5$; samples per window per box $K_b=4$}
\KwResult{Per-fold accuracy, per-class F1, confusion matrices}

\tcp{Phase 1: parse records}
Initialise empty table $\mathcal{R}$\;
\ForEach{image file $p$ in \texttt{DATA\_DIR}}{
  parse $(\textit{date}, \textit{box}, \textit{modality})$ from filename of $p$\;
  append $\langle p,\ \textit{class},\ \textit{date},\ \textit{box}\rangle$ to $\mathcal{R}$\;
}
\textbf{assert} each box maps to exactly one class \tcp*{class--box confound is by design}

\tcp{Phase 2: build per-box temporal sequences}
Initialise sequence pool $\mathcal{S}\leftarrow\varnothing$\;
\ForEach{$(\textit{class}, \textit{box})$ group in $\mathcal{R}$}{
  $D \leftarrow$ sorted unique dates of this box\;
  \For{$i \leftarrow 0$ \KwTo $|D| - T$}{
    $W \leftarrow D[i \,..\, i{+}T{-}1]$ \tcp*{window of $T$ consecutive dates}
    \For{$k \leftarrow 1$ \KwTo $K_b$}{
      $s \leftarrow$ one image sampled per date in $W$ (modality collapsed)\;
      add $s$ to $\mathcal{S}$ with label \textit{class} and tag \textit{box}\;
    }
  }
}

\tcp{Phase 3: define leave-one-box-out folds (stratified within class)}
\ForEach{$\textit{class}\ c$}{
  $B_c \leftarrow$ sorted list of boxes belonging to $c$\;
}
$F \leftarrow \min_c |B_c|$ \tcp*{number of folds $=$ boxes per class $= 9$}

\tcp{Phase 4: box-disjoint CV loop}
\For{$f \leftarrow 1$ \KwTo $F$}{
  $\mathcal{T}_{\text{test}} \leftarrow \{\, B_c[f] : \text{for each class } c \,\}$ \tcp*{one held-out box per class}
  $\textit{test} \leftarrow \{\, s \in \mathcal{S} : \textit{box}(s) \in \mathcal{T}_{\text{test}} \,\}$\;
  $\textit{train} \leftarrow \{\, s \in \mathcal{S} : \textit{box}(s) \notin \mathcal{T}_{\text{test}} \,\}$\;

  \textbf{assert} $\text{boxes}(\textit{train}) \cap \text{boxes}(\textit{test}) = \varnothing$ \tcp*{identity disjoint}
  \textbf{assert} $\text{files}(\textit{train}) \cap \text{files}(\textit{test}) = \varnothing$ \tcp*{no shared image}
  \lIf{$|\textit{train}| < \tau$ \textbf{or} $|\textit{test}| = 0$}{skip fold}

  augment \textit{train} only ($2\times$, same transform across all $T$ frames)\;
  initialise a fresh CNN--LSTM model (MobileNetV2 pretrained on ImageNet; fine-tune the final 10 layers)\;
  train on \textit{train}; evaluate on held-out \textit{test}\;
  record fold accuracy, per-class F1, and confusion matrix\;
}
\end{algorithm}

\subsection{Spatial Baseline Framework}
The design of the CNN--LSTM framework began with the search for a lightweight spatial feature extractor with fewer parameters. In our experiments, three lightweight CNN models, namely MobileNetV2, EfficientNetB0, and NASNetMobile, were explored for distinguishing nitrogen severity classes. Among these models, MobileNetV2 achieved the best performance and was therefore selected as the feature extractor in the proposed CNN--LSTM framework. 
For the spatial-only setup, we employed pretrained \textit{MobileNetV2} with weights initialized from the \textit{ImageNet} dataset. The original top layer, configured for 1,000 ImageNet classes, was removed so the backbone could function as a feature extractor. Custom classification layers were appended to adapt the model for our three-class classification task. By leveraging pretrained weights, we utilized the rich feature representations learned from large-scale data while fine-tuning the model to our target domain.

\begin{figure*}[!t]
        \centering
		\includegraphics[scale=0.95]{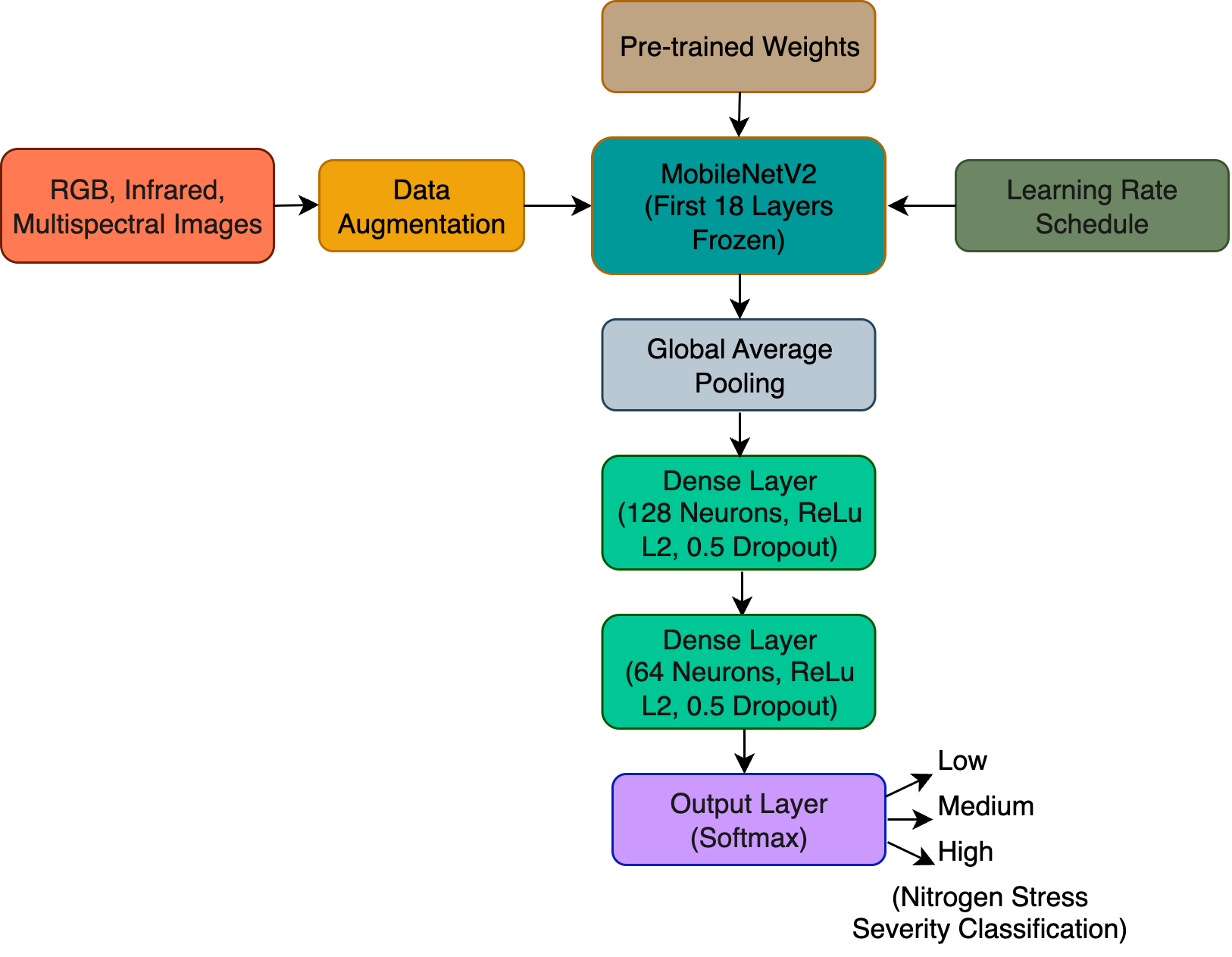}
		\caption{Spatial Deep Learning Framework for Nitrogen Stress Severity Classification} \label{frame_spatial}
\end{figure*}

To retain essential feature extraction capabilities, the first 18 layers of \textit{MobileNetV2} were frozen, while the subsequent layers were fine-tuned. On top of the backbone, a \textit{GlobalAveragePooling2D} layer reduced spatial dimensions, followed by two dense layers (128 and 64 neurons) with ReLU activation and L2 regularization. Dropout layers with a rate of 0.5 were added after each dense layer to improve generalization. The final classification layer used softmax activation to predict probabilities across the three categories. The architecture is illustrated in Fig. \ref{frame_spatial}. 

To improve model performance and address limited training data, extensive data augmentation is performed using random rotations, shear transformations, horizontal and vertical flips, and spatial translations. The model is trained using the \textit{Adam} optimizer with an exponentially decaying learning rate and evaluated under a 5-fold stratified cross-validation protocol. 

\section{Results and Discussion}
The spatio-temporal and spatial-only frameworks were implemented in Python (version 3.10.14) using machine learning libraries, including \textit{Keras}, \textit{TensorFlow}, \textit{Scikit-learn}, \textit{Pandas}, \textit{NumPy}, and \textit{Matplotlib}.

\subsection{Performance Evaluation of Spatial Temporal Framework with Forward Chain Cross Validation}

The CNN--LSTM framework was evaluated using the forward-chaining cross-validation strategy described in Algorithm \ref{algo:cnn_lstm}. Unlike random $k$-fold cross-validation, this strategy preserves the chronological order of the image sequences and prevents future data from entering the training phase. The proposed framework was tested with different subsets of parameters for 60 epochs. The best performance was obtained using the parameter settings summarized in Table \ref{tab:lstm_params}.

\begin{table}[ht]
\centering
\caption{Training and data-augmentation configuration of the CNN--LSTM framework}
\label{tab:lstm_params}
\begin{tabular}{|p{5.5cm}|p{8cm}|}
\hline
\multicolumn{2}{|c|}{\textbf{Training Parameters}} \\
\hline
\textbf{Parameter} & \textbf{Value} \\
\hline
Image Size & $224 \times 224$ pixels \\
Batch Size & 16 \\
Sequence Length & 5 \\
Samples per Window per Class & 36 \\
Initial Learning Rate & $1 \times 10^{-4}$ with an exponential decay factor of 0.9 every 1000 steps \\
Maximum Epochs & 60 \\
Early Stopping & Patience = 10, monitored on validation loss \\
Best Model Selection & Minimum validation loss \\
\hline
\multicolumn{2}{|c|}{\textbf{Augmentation Parameters}} \\
\hline
Shear Range & 0.2 \\
Rotation Range & $15^{\circ}$ \\
Width Shift Range & 0.1 \\
Height Shift Range & 0.1 \\
Horizontal Flip & Enabled \\
Vertical Flip & Enabled \\
Fill Mode & Nearest \\
\hline
\end{tabular}
\end{table}

In each fold, all sequences whose terminal date, i.e., \textit{last\_date}, occurred before the validation anchor were assigned to the training set, whereas sequences whose \textit{last\_date} matched the validation anchor were used for validation/evaluation. Thus, the validation set remained fixed at 108 sequences per fold ($1$ window $\times$ 36 samples $\times$ 3 classes), while the training set increased monotonically as the validation anchor advanced. After sequence-level augmentation, the effective training set sizes increased from 864 sequences in Fold~1 to 1944 sequences in Fold~6. Therefore, the reported results represent temporally ordered generalization to future acquisition dates rather than random sample-level generalization.

As shown in Table~\ref{tab:fold_results}, the CNN--LSTM framework achieved high and stable performance across the six forward-chaining folds. The mean training accuracy was $0.9862 \pm 0.0113$, while the mean held-out validation/evaluation accuracy was $0.9537 \pm 0.0142$. Fold~4 produced the highest validation/evaluation accuracy of $0.9815$, whereas the remaining folds ranged from $0.9444$ to $0.9537$. The best checkpoints occurred between epochs 27 and 49, indicating that the model generally converged before the maximum training limit. Since the held-out forward-chain anchor was used for fold-wise evaluation, the validation and test accuracy columns in Table~\ref{tab:fold_results} are identical.

\begin{table*}[htbp]
\centering
\caption{Fold-wise best performance metrics of the CNN--LSTM spatio-temporal framework during 6-fold cross-validation.}
\label{tab:fold_results}
\begin{tabular}{c c c c c c}
\hline
\textbf{Fold} & \textbf{Train Accuracy} & \textbf{Train Loss} &
\textbf{Val Accuracy} & \textbf{Val Loss} & \textbf{Epoch} \\
\hline
1 & 0.9988 & 0.0752 & 0.9444 & 0.1552 & 49 \\
2 & 0.9676 & 0.1459 & 0.9537 & 0.2129 & 27 \\
3 & 0.9954 & 0.0748 & 0.9537 & 0.1733 & 48 \\
4 & 0.9854 & 0.1152 & 0.9815 & 0.1043 & 32 \\
5 & 0.9815 & 0.1151 & 0.9444 & 0.1822 & 33 \\
6 & 0.9882 & 0.0977 & 0.9444 & 0.2023 & 31 \\
\hline
\textbf{Mean} & 0.9862 & 0.1040 & 0.9537 & 0.1717 & -- \\
\textbf{Std}  & 0.0113 & 0.0270 & 0.0142 & 0.0392 & -- \\
\hline
\end{tabular}
\end{table*}

\begin{figure*}[hbtp]
    \centering

    \begin{subfigure}[b]{0.32\linewidth}
        \centering
        \includegraphics[width=\linewidth]{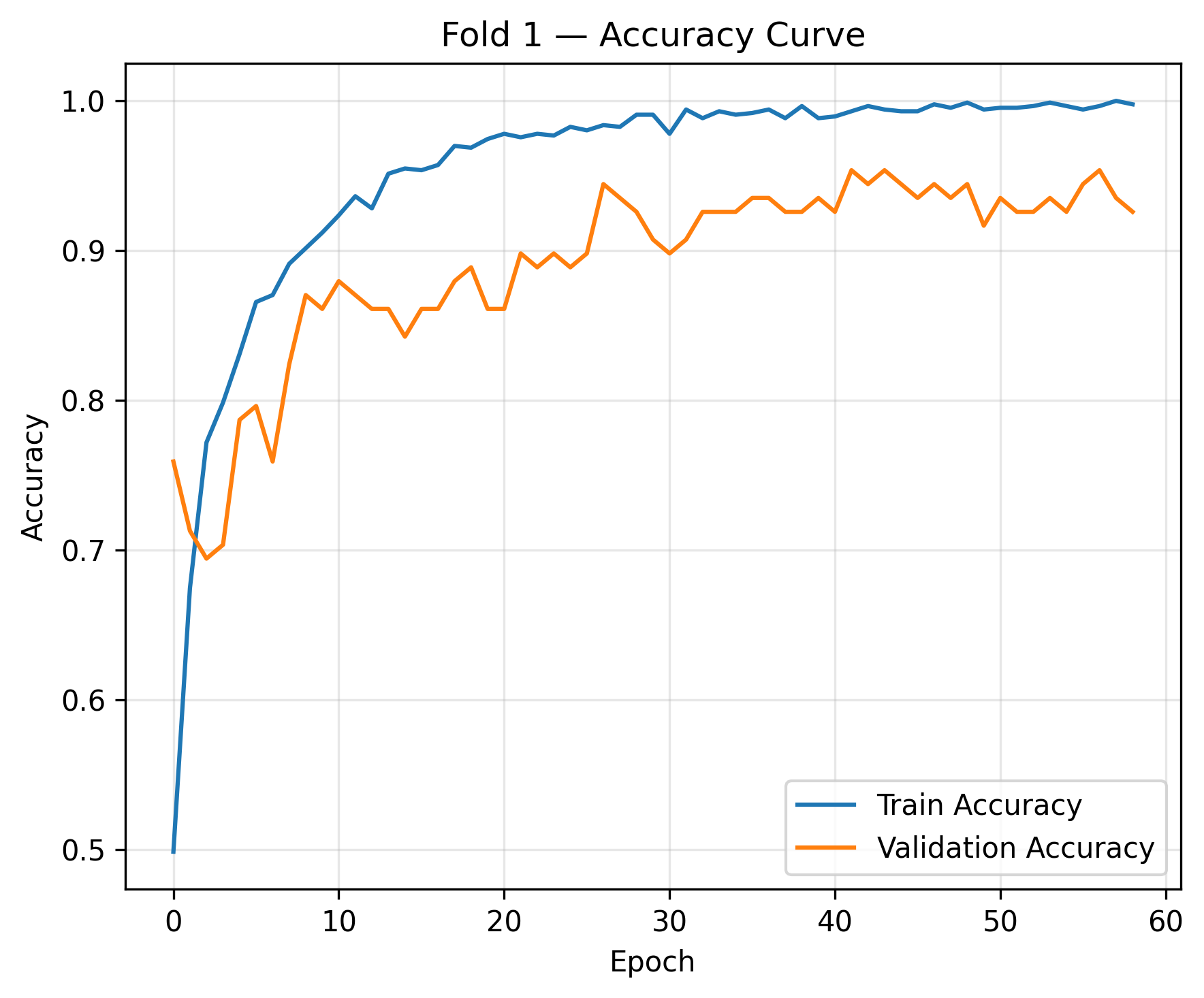}
        \caption{Fold 1}
        \label{fig:fold1_acc_st}
    \end{subfigure}
    \begin{subfigure}[b]{0.32\linewidth}
        \centering
        \includegraphics[width=\linewidth]{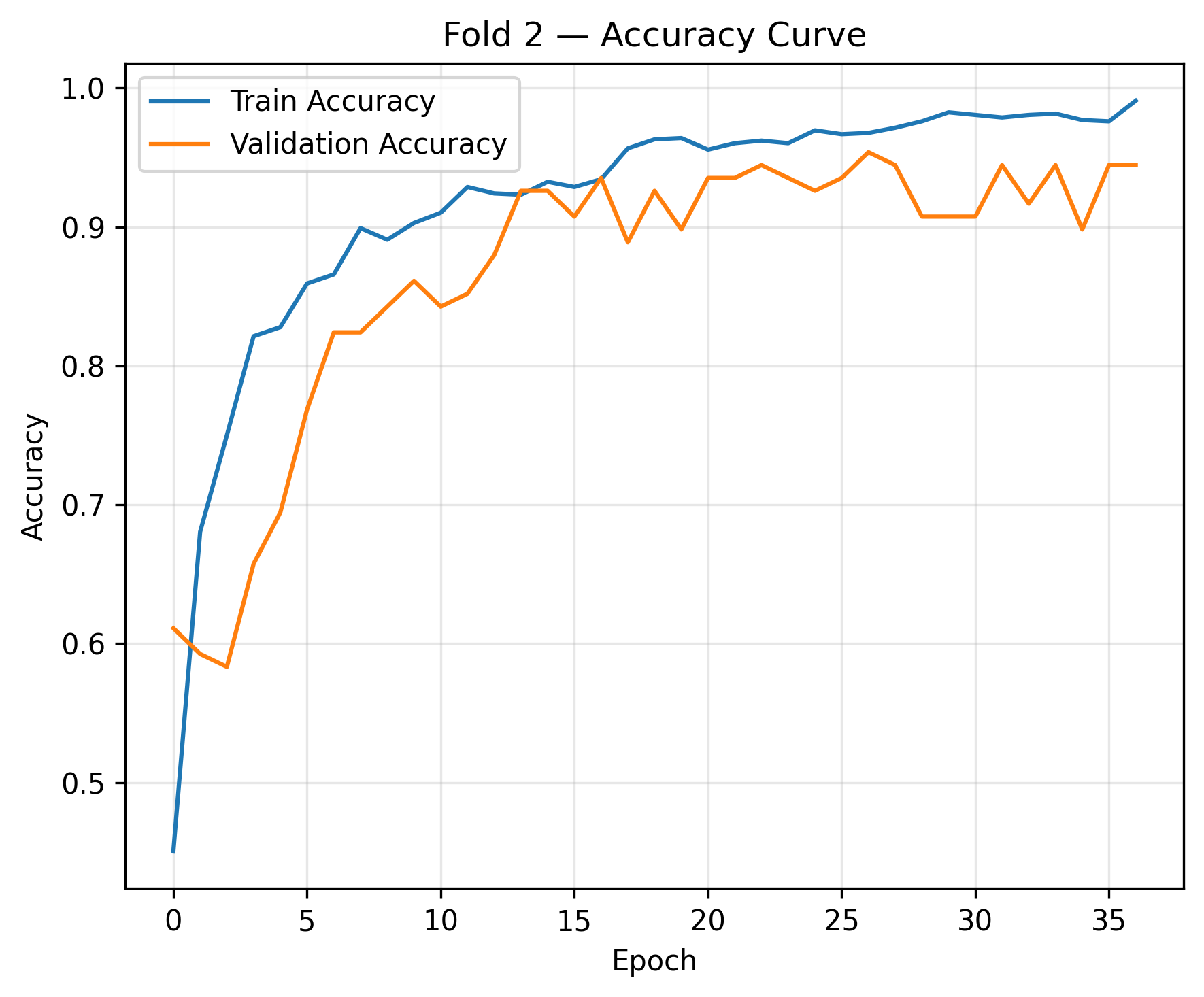}
        \caption{Fold 2}
        \label{fig:fold2_acc_st}
    \end{subfigure}
    \begin{subfigure}[b]{0.32\linewidth}
        \centering
        \includegraphics[width=\linewidth]{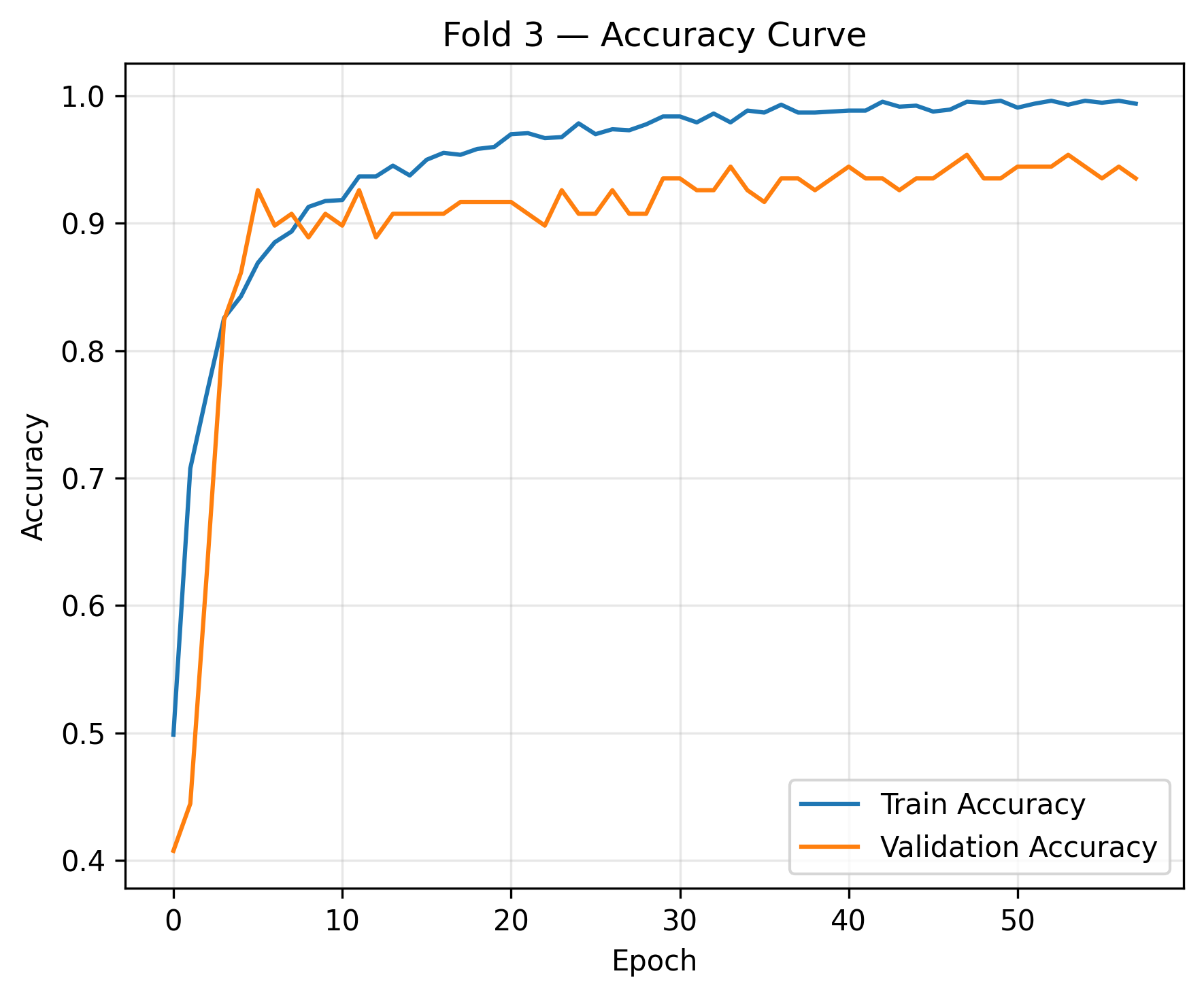}
        \caption{Fold 3}
        \label{fig:fold3_acc_st}
    \end{subfigure}

    \vspace{0.7em}

    \begin{subfigure}[b]{0.32\linewidth}
        \centering
        \includegraphics[width=\linewidth]{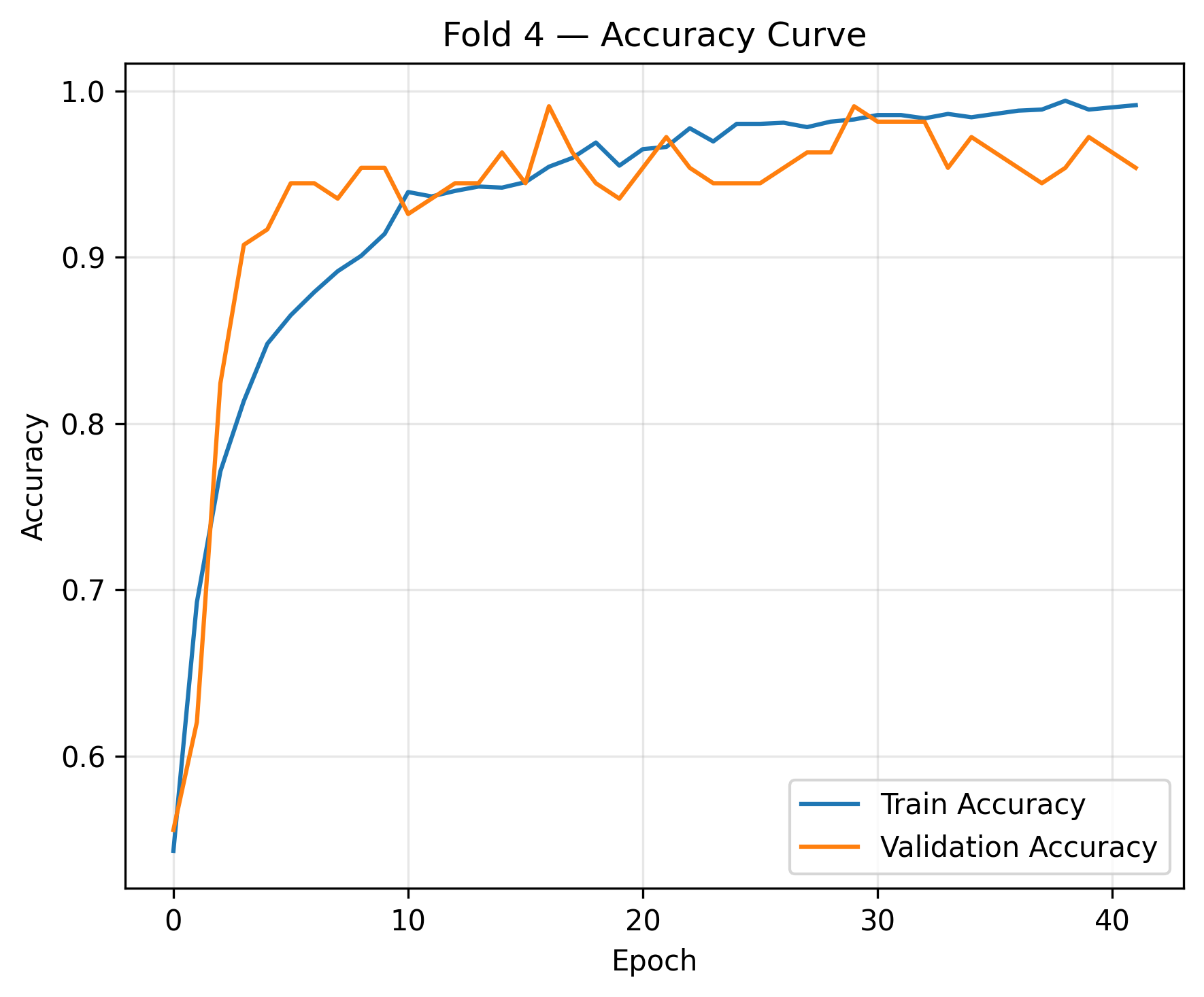}
        \caption{Fold 4}
        \label{fig:fold4_acc_st}
    \end{subfigure}
    \begin{subfigure}[b]{0.32\linewidth}
        \centering
        \includegraphics[width=\linewidth]{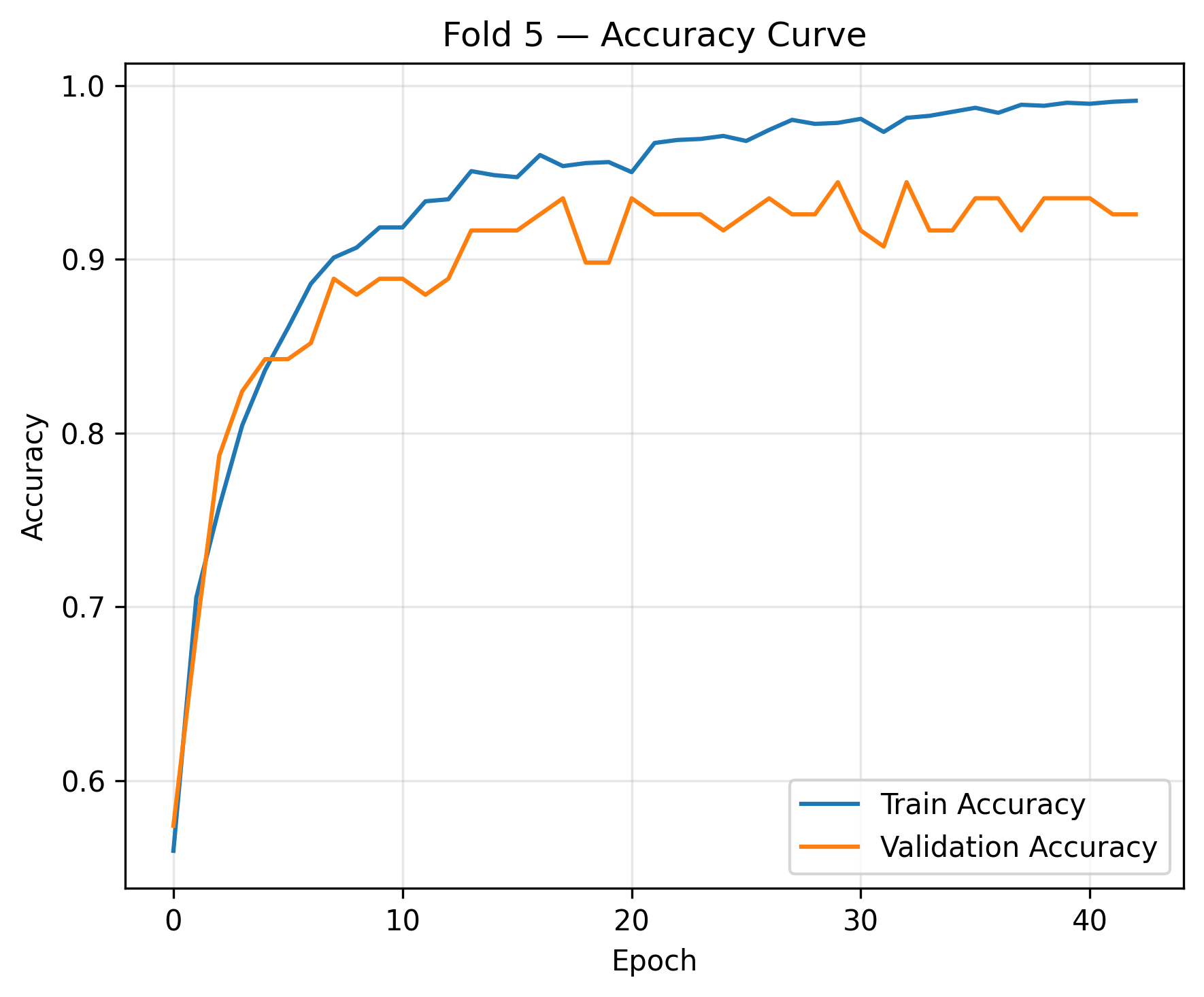}
        \caption{Fold 5}
        \label{fig:fold5_acc_st}
    \end{subfigure}
     \begin{subfigure}[b]{0.32\linewidth}
        \centering
        \includegraphics[width=\linewidth]{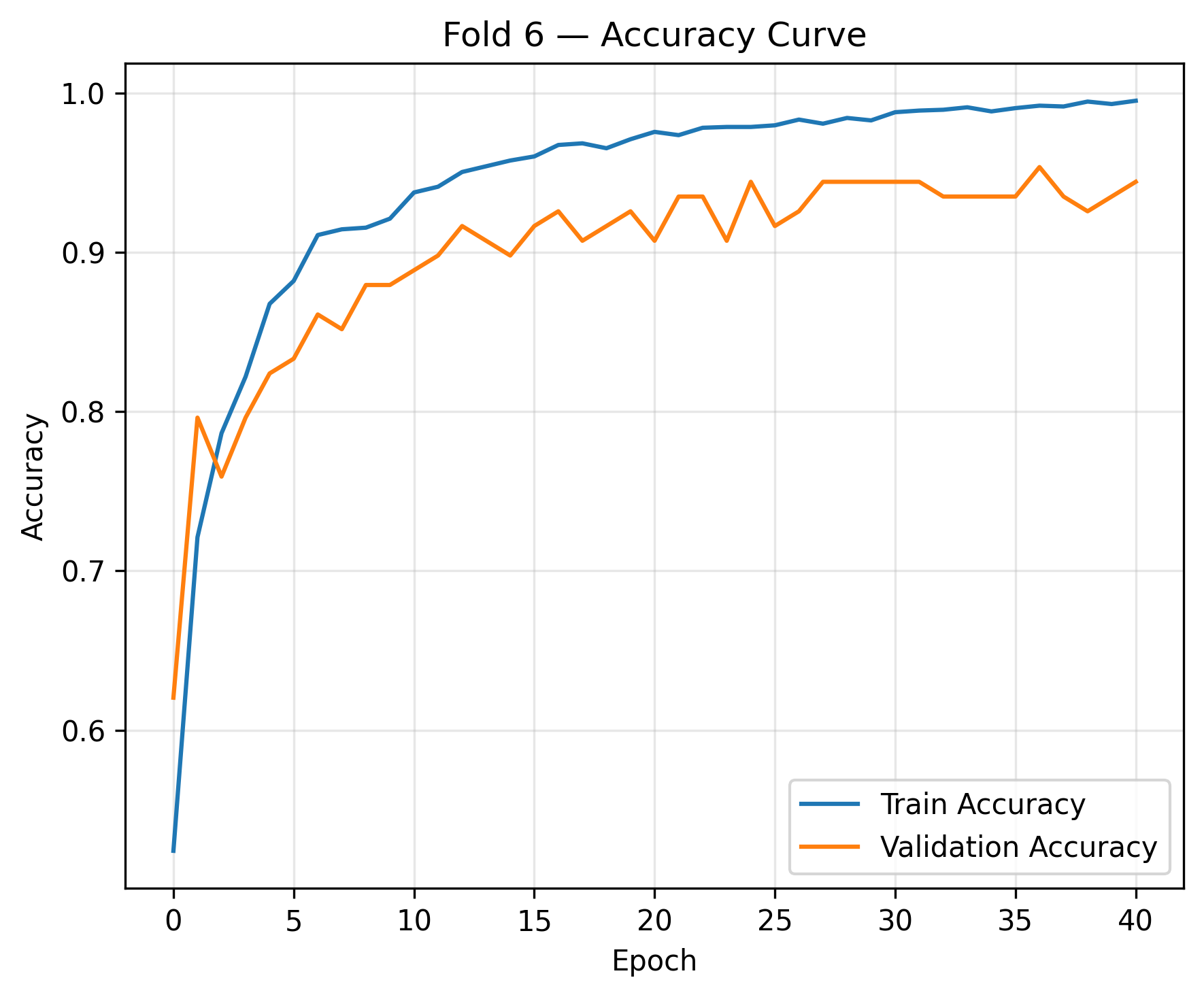}
        \caption{Fold 6}
        \label{fig:fold6_acc_st}
    \end{subfigure}

    \caption{Accuracy curves of CNN-LSTM for forward chain cross-validation (a–f correspond to Fold 1–6).}
    \label{fig:mobilenetv2_lstm_acc}
\end{figure*}

\begin{figure*}[h!]
    \centering

    \begin{subfigure}[b]{0.32\linewidth}
        \centering
        \includegraphics[width=\linewidth]{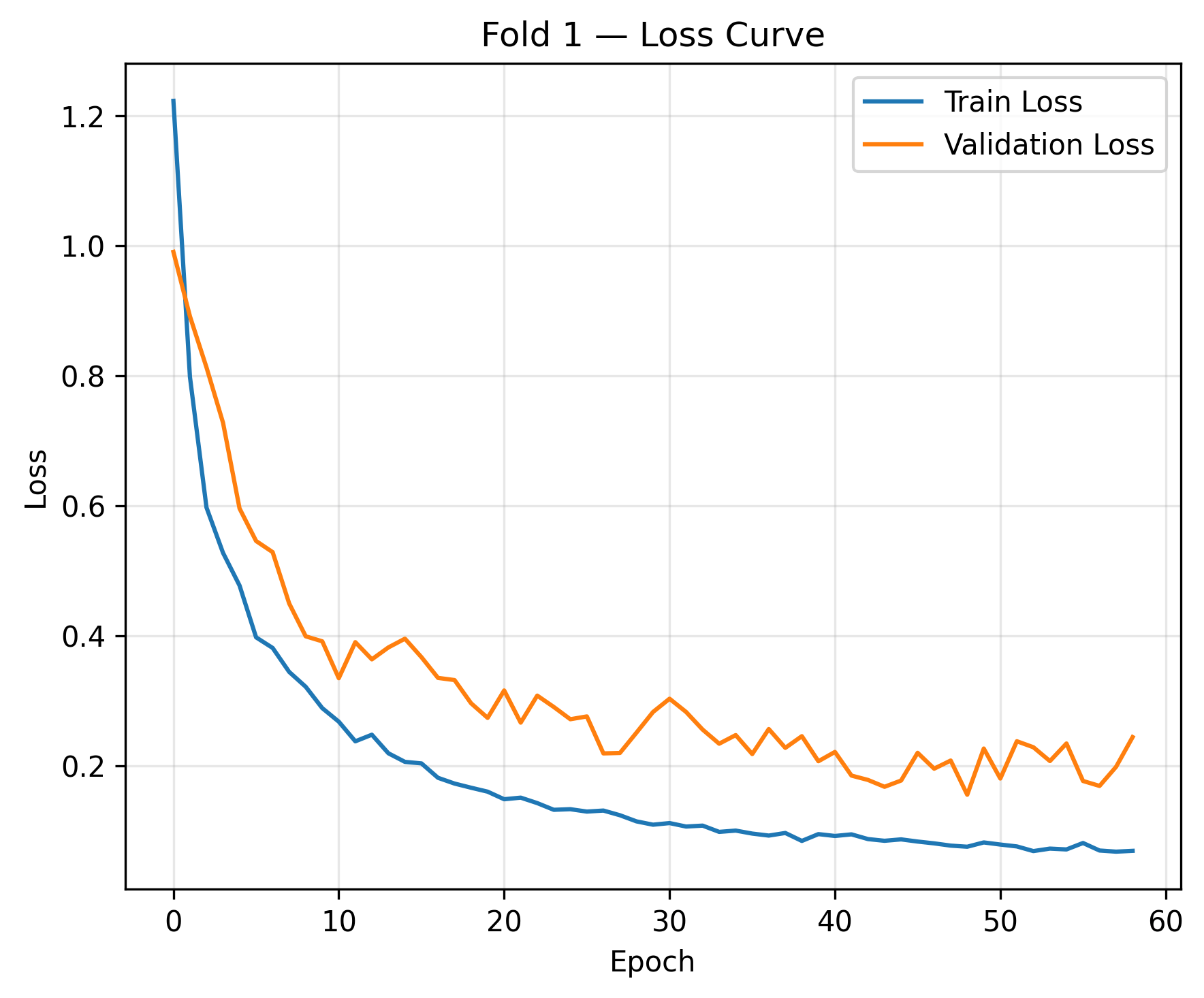}
        \caption{Fold 1}
        \label{fig:fold1_loss_st}
    \end{subfigure}
    \begin{subfigure}[b]{0.32\linewidth}
        \centering
        \includegraphics[width=\linewidth]{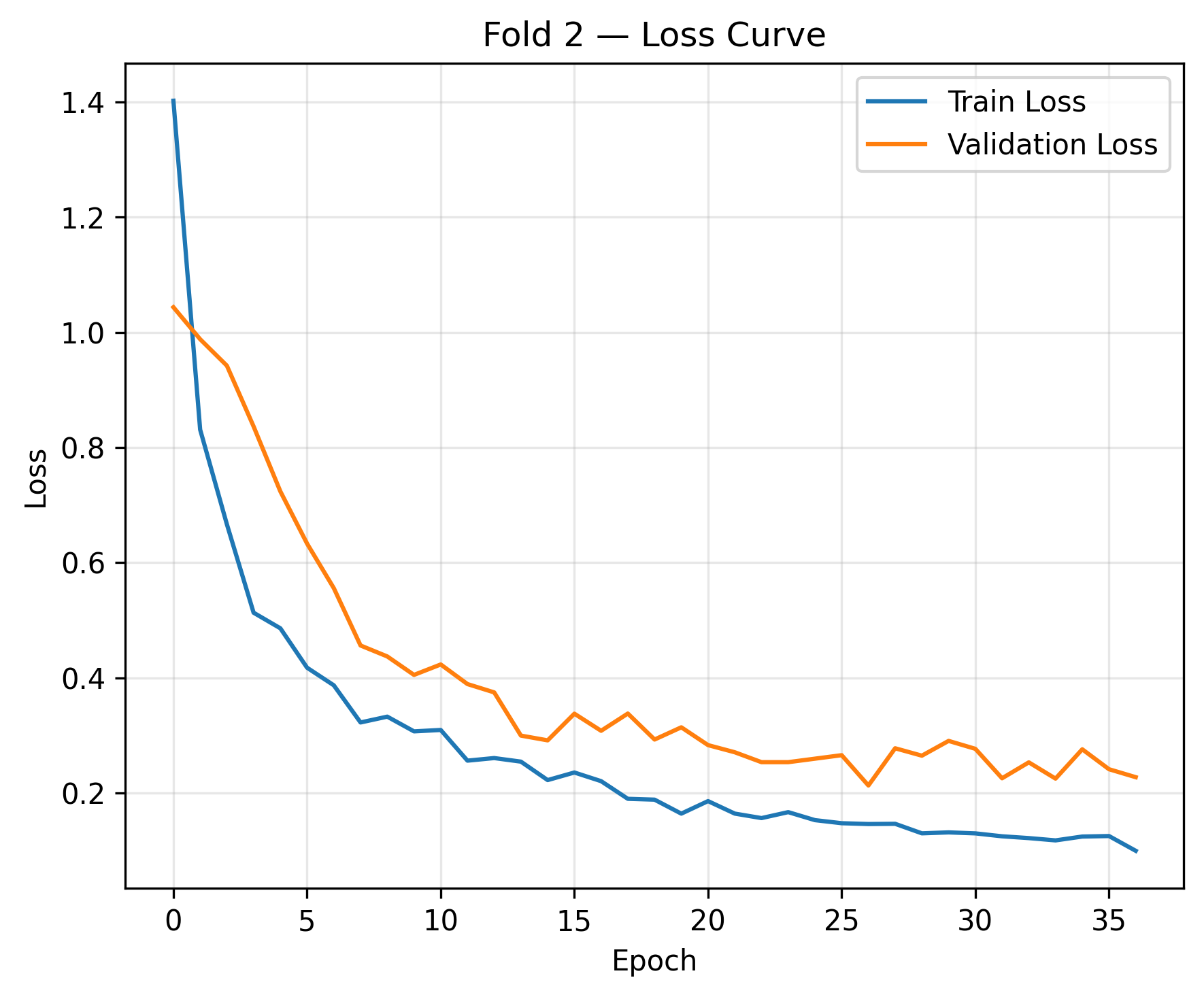}
        \caption{Fold 2}
        \label{fig:fold2_loss_st}
    \end{subfigure}
    \begin{subfigure}[b]{0.32\linewidth}
        \centering
        \includegraphics[width=\linewidth]{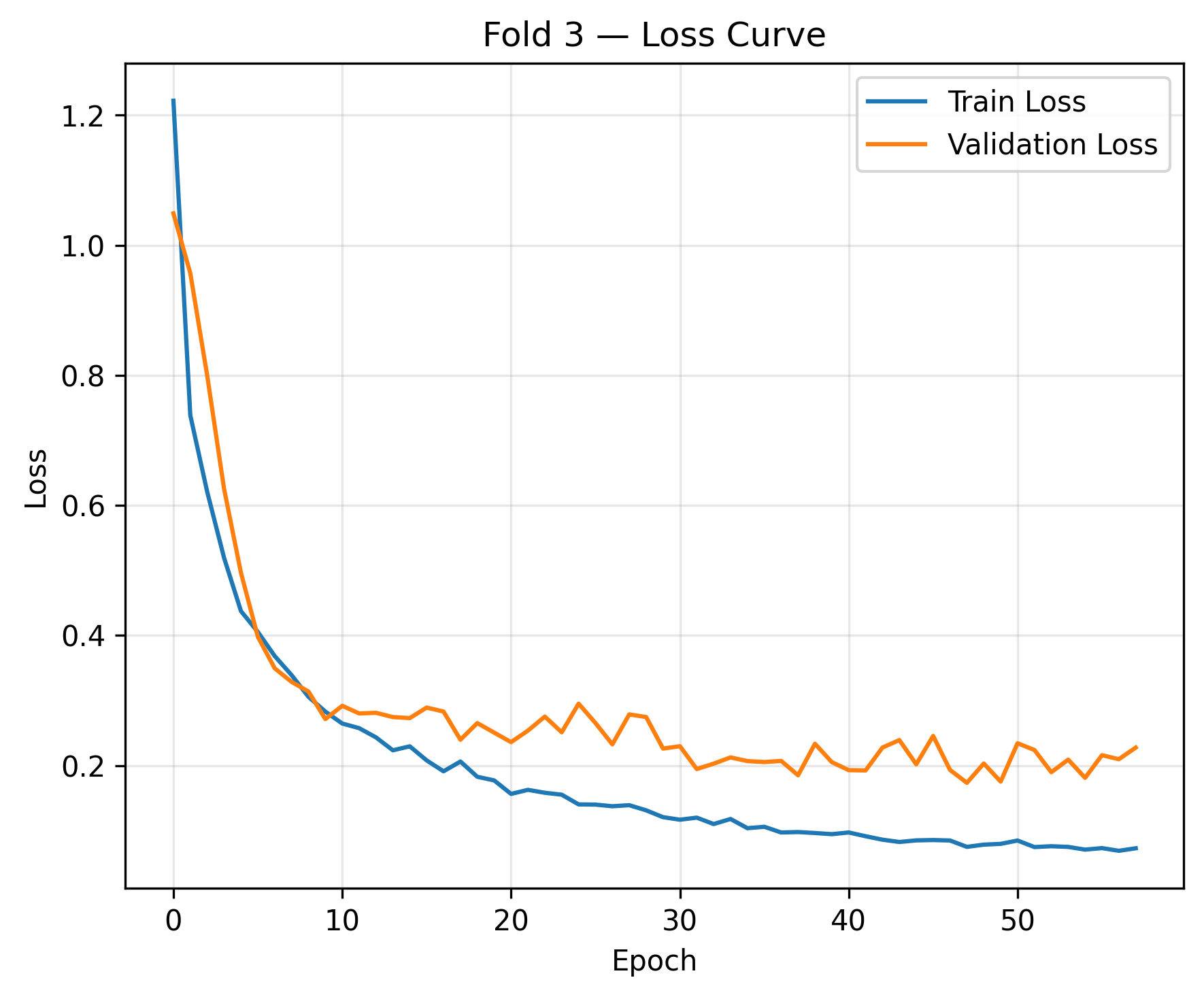}
        \caption{Fold 3}
        \label{fig:fold3_loss_st}
    \end{subfigure}

    \vspace{0.7em}

    \begin{subfigure}[b]{0.32\linewidth}
        \centering
        \includegraphics[width=\linewidth]{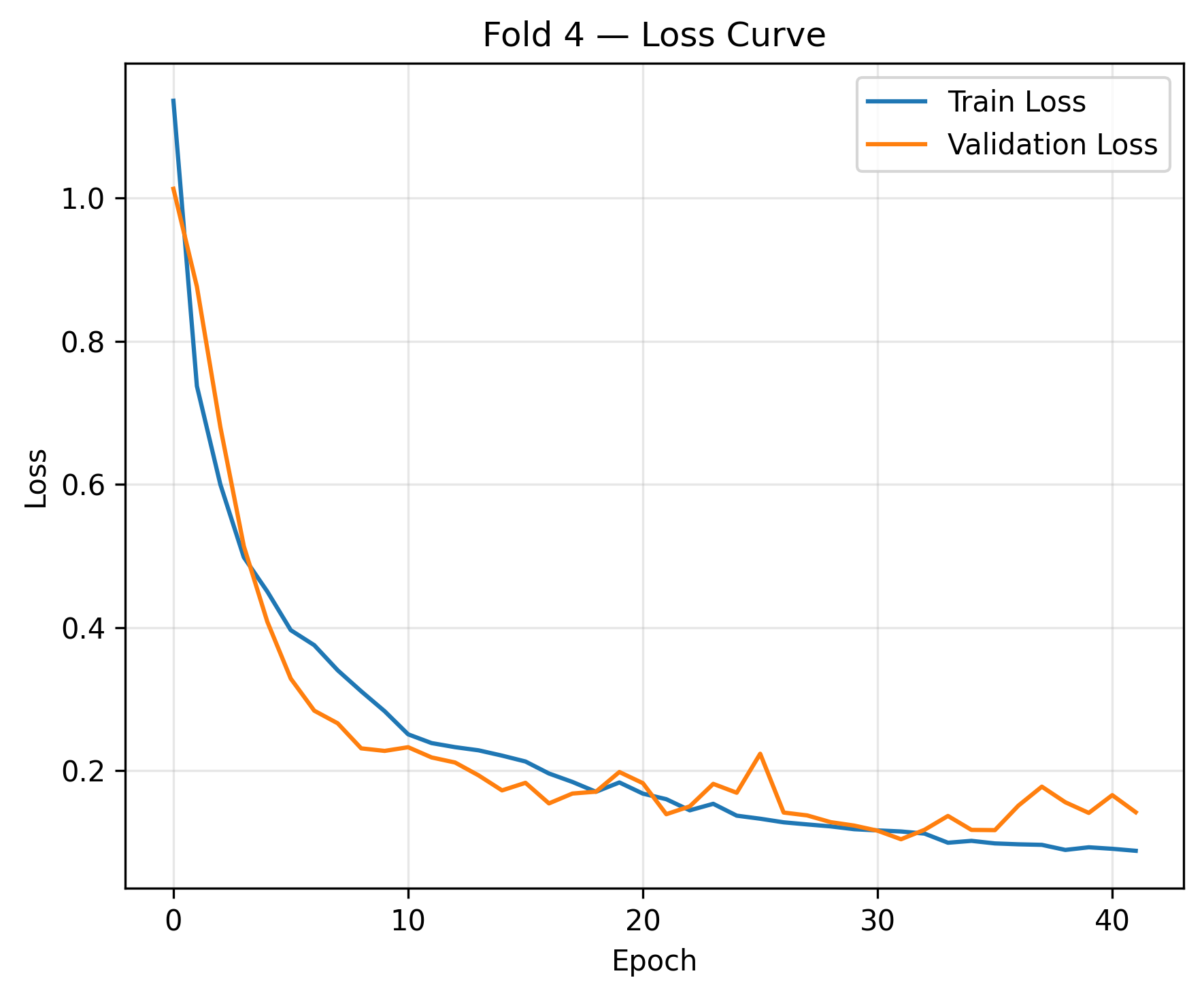}
        \caption{Fold 4}
        \label{fig:fold4_loss_st}
    \end{subfigure}
    \begin{subfigure}[b]{0.32\linewidth}
        \centering
        \includegraphics[width=\linewidth]{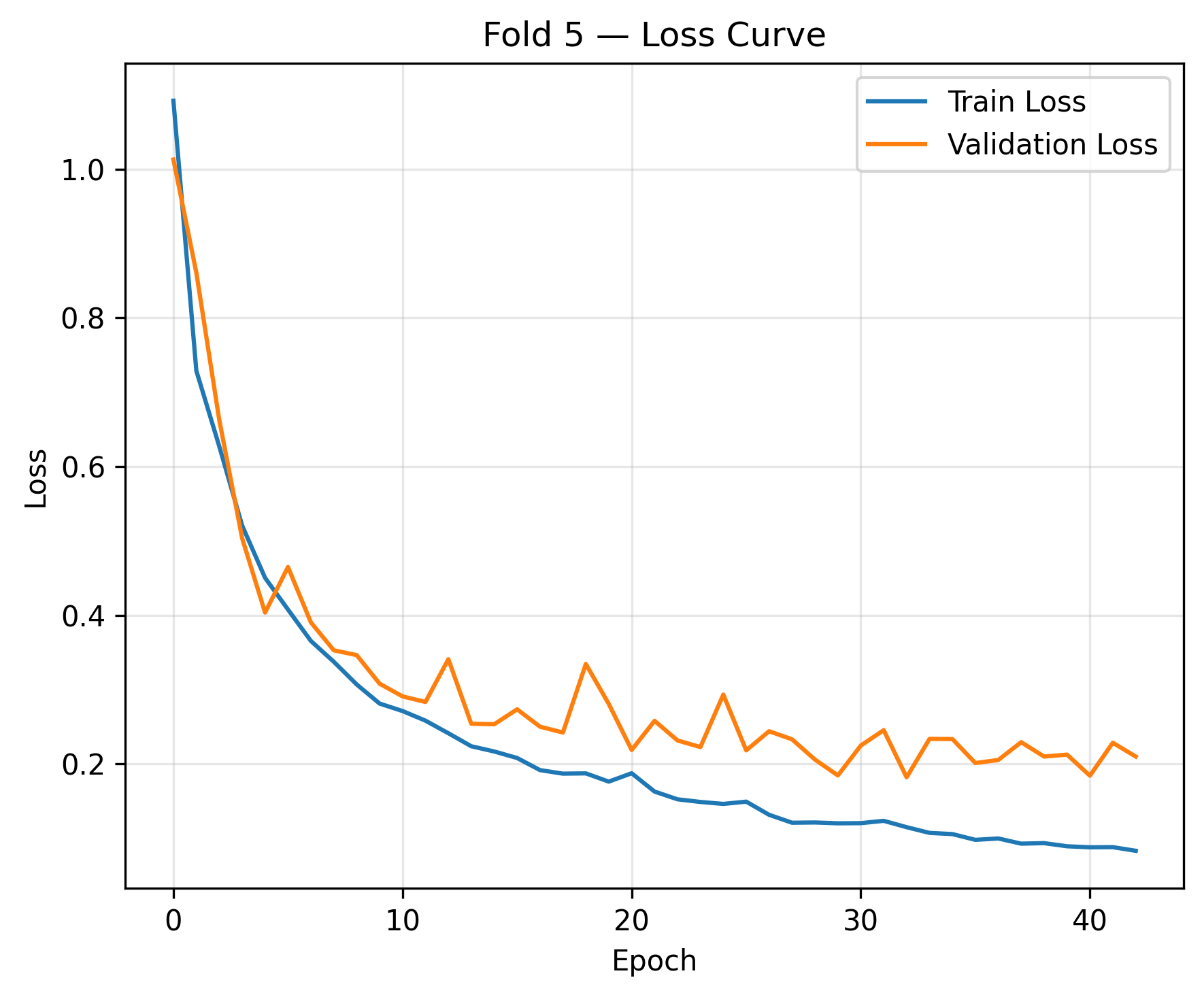}
        \caption{Fold 5}
        \label{fig:fold5_loss_st}
    \end{subfigure}
    \begin{subfigure}[b]{0.32\linewidth}
        \centering
        \includegraphics[width=\linewidth]{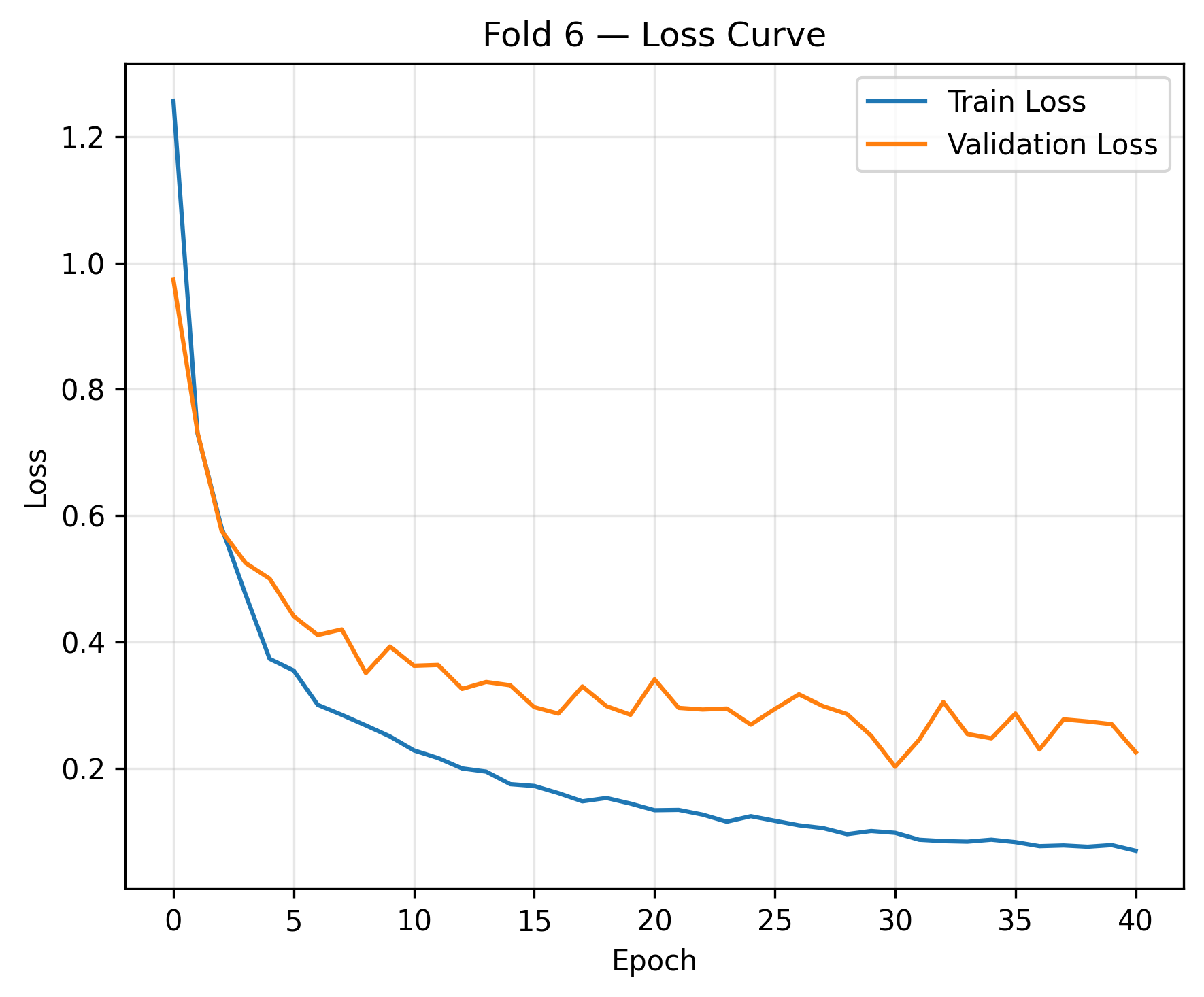}
        \caption{Fold 6}
        \label{fig:fold6_loss_st}
    \end{subfigure}

    \caption{Loss curves of CNN-LSTM for forward chain cross-validation (a–f correspond to Fold 1–6).}
    \label{fig:mobilenetv2_lstm_loss}
\end{figure*}

Fig.~\ref{fig:mobilenetv2_lstm_acc} shows that the model converged rapidly in all six folds, with validation accuracy stabilizing at high levels after the initial epochs. However, the curves also indicate a moderate training--validation gap in some folds, particularly Folds~1, 5, and 6, suggesting that the model learned strong temporal representations while still facing some fold-specific generalization difficulty. The loss curves in Fig.~\ref{fig:mobilenetv2_lstm_loss} further support this observation: training loss declined consistently, while validation loss remained stable without evidence of uncontrolled divergence. The lowest validation loss was observed in Fold~4, which also achieved the highest validation/evaluation accuracy.

\begin{figure*}[hbtp]
    \centering

    \begin{subfigure}[b]{0.32\linewidth}
        \centering
        \includegraphics[width=\linewidth]{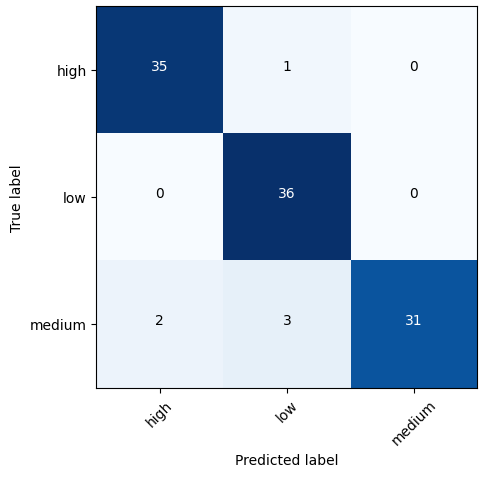}
        \caption{Fold 1}
        \label{fig:fold1_conf}
    \end{subfigure}
    \begin{subfigure}[b]{0.32\linewidth}
        \centering
        \includegraphics[width=\linewidth]{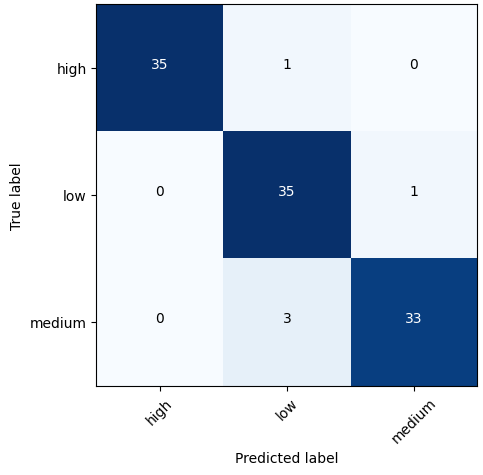}
        \caption{Fold 2}
        \label{fig:fold2_conf}
    \end{subfigure}
    \begin{subfigure}[b]{0.32\linewidth}
        \centering
        \includegraphics[width=\linewidth]{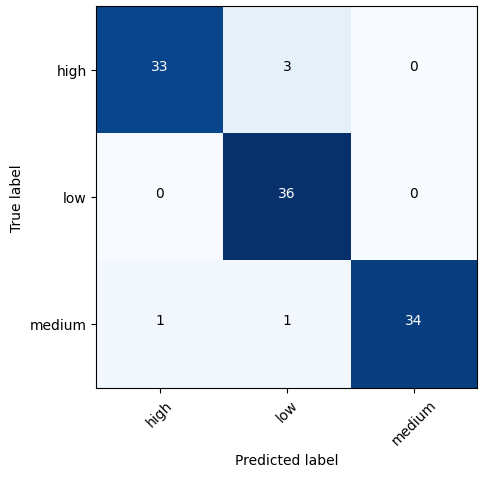}
        \caption{Fold 3}
        \label{fig:fold3_conf}
    \end{subfigure}

    \vspace{0.7em}

    \begin{subfigure}[b]{0.32\linewidth}
        \centering
        \includegraphics[width=\linewidth]{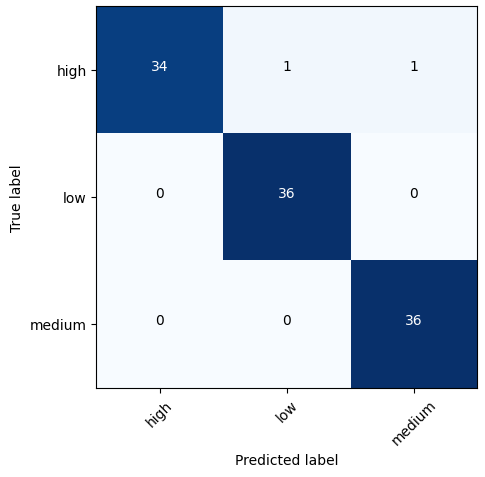}
        \caption{Fold 4}
        \label{fig:fold4_conf}
    \end{subfigure}
    \begin{subfigure}[b]{0.32\linewidth}
        \centering
        \includegraphics[width=\linewidth]{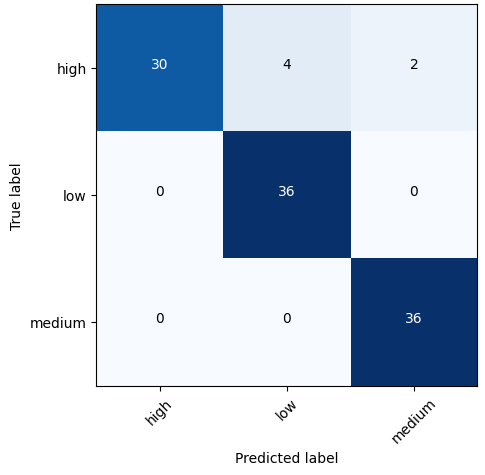}
        \caption{Fold 5}
        \label{fig:fold5_conf}
    \end{subfigure}
    \begin{subfigure}[b]{0.32\linewidth}
        \centering
        \includegraphics[width=\linewidth]{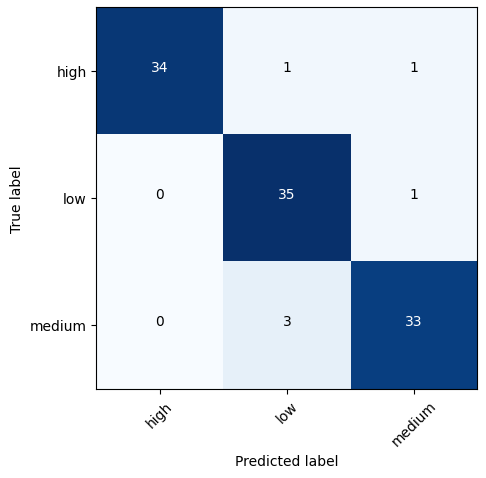}
        \caption{Fold 6}
        \label{fig:fold6_conf}
    \end{subfigure}

    \caption{Confusion Matrices of CNN-LSTM for forward chain cross-validation validation (a–f correspond to Fold 1–6.}
    \label{fig:mobilenetv2_lstm_conf}
\end{figure*}

The confusion matrices in Fig.~\ref{fig:mobilenetv2_lstm_conf} show that most samples were correctly classified across the three nitrogen stress categories. Fold~4 showed the strongest class separation, with perfect classification of the low and medium nitrogen classes and only limited misclassification in the high nitrogen class. In contrast, Fold~5 showed the largest difficulty for the high nitrogen class, where 30 of 36 samples were correctly classified and the remaining samples were confused with low or medium classes. Overall, the diagonal dominance of the confusion matrices confirms that the framework classified the treatment labels with high sequence-level validation accuracy under this evaluation setting.

Table~\ref{tab:prf_folds_cnn_lstm} further confirms the class-wise robustness of the model, although the metrics were not uniformly above 0.97 in every fold. The low nitrogen class showed consistently high recall values ranging from 0.97 to 1.00, indicating strong sensitivity for detecting nitrogen-deficient samples. The high nitrogen class achieved excellent precision across folds, ranging from 0.95 to 1.00, although its recall decreased to 0.83 in Fold~5. The medium nitrogen class also showed strong performance, with F1-scores ranging from 0.93 to 0.99. Across all classes and folds, the F1-scores ranged from 0.91 to 0.99, and the average macro-F1 score was approximately 0.954, consistent with the mean fold-wise accuracy.
\begin{table*}[htbp]
\centering
\caption{Class-wise precision, recall, F1-score, and accuracy across six CNN-LSTM cross-validation folds.}
\label{tab:prf_folds_cnn_lstm}
\begin{tabular}{lccccccc}
\hline
\textbf{Class} & \textbf{Metric} & \textbf{Fold 1} & \textbf{Fold 2} & \textbf{Fold 3} & \textbf{Fold 4} & \textbf{Fold 5} & \textbf{Fold 6} \\
\hline

\multirow{3}{*}{High} 
 & Precision & 0.95 & 1.00 & 0.97 & 1.00 & 1.00 & 1.00 \\
 & Recall    & 0.97 & 0.97 & 0.92 & 0.94 & 0.83 & 0.94 \\
 & F1-score  & 0.96 & 0.99 & 0.94 & 0.97 & 0.91 & 0.97 \\
\hline

\multirow{3}{*}{Low} 
 & Precision & 0.90 & 0.90 & 0.90 & 0.97 & 0.90 & 0.90 \\
 & Recall    & 1.00 & 0.97 & 1.00 & 1.00 & 1.00 & 0.97 \\
 & F1-score  & 0.95 & 0.93 & 0.95 & 0.99 & 0.95 & 0.93 \\
\hline

\multirow{3}{*}{Medium} 
 & Precision & 1.00 & 0.97 & 1.00 & 0.97 & 0.95 & 0.94 \\
 & Recall    & 0.86 & 0.92 & 0.94 & 1.00 & 1.00 & 0.92 \\
 & F1-score  & 0.93 & 0.94 & 0.97 & 0.99 & 0.97 & 0.93 \\
\hline

\multicolumn{2}{l}{\textbf{Accuracy}} 
& 0.94 & 0.95 & 0.95 & 0.98 & 0.94 & 0.94 \\
\hline

\end{tabular}
\end{table*}

To provide a biological interpretation of
the temporally ordered evaluation, the acquisition dates were grouped into
growth-stage intervals. In the forward-chaining protocol, each validation fold
was anchored by the terminal date of the validation sequence, i.e.,
\textit{last\_date}. Therefore, the fold-wise accuracies can be interpreted as
growth-stage-dependent performance estimates, while noting that each validation
sample still contains five consecutive acquisition dates and therefore includes
temporal context from the preceding four dates.

\begin{table}[htbp]
\centering
\caption{Growth-stage mapping of retained acquisition dates and forward-chaining validation folds.}
\label{tab:growth_stage_mapping}
\begin{tabular}{p{3.2cm} p{3.6cm} p{4.8cm} p{2.5cm}}
\hline
\textbf{Retained dates} & \textbf{Calendar period} & \textbf{Interpreted growth stage} & \textbf{Validation fold}\\
\hline
D1--D4 & 1--12 February & Early vegetative stage; used as temporal context in training sequences & None \\
D5--D8 & 15--26 February & Mid-vegetative stage; contributes mainly as contextual frames in early validation sequences & Folds~1--2 \\
D9--D11 & 2--8 March & Late vegetative stage with more visible canopy-level stress expression & Folds~3--4 \\
D12--D14 & 12--29 March & Rapid growth phase approaching bolting, with more established stress phenotypes & Folds~5--6 \\
\hline
\end{tabular}
\end{table}

As shown in Table~\ref{tab:growth_stage_mapping}, the first four dates
(D1--D4) provided early temporal context and were not used as validation
anchors. Dates
D5--D8 correspond to the mid-vegetative stage, where early stress-related
changes such as reduced leaf expansion and initial chlorosis may begin to
emerge; this stage is represented by Folds~1--2. Dates D9--D11 correspond to
the late vegetative stage, during which canopy development increases and
stress-related differences become more visually distinguishable; this period is
represented by Folds~3--4. Finally, Dates D12--D14 represent the rapid growth
phase approaching bolting, where stress phenotypes are more established; this
stage is represented by Folds~5--6. The forward-chaining cross-validation design provides not only a
temporally ordered evaluation but also a growth-stage-wise interpretation of
model performance. The fold-wise accuracies for
Folds~1--6 were 0.9444, 0.9537, 0.9537, 0.9815, 0.9444, and 0.9444,
respectively. These results indicate consistently high performance across the
growth stages. These results indicate that the CNN--LSTM framework maintained consistently high sequence-level validation accuracy across the evaluated acquisition-date anchors.

While Khanna et al.\cite{khanna_spatio_2019} utilized the same dataset, leveraging vegetation indices, hyperspectral signatures, and 3D point cloud features over a two-month crop cycle, their modeling approach lacked dynamic learning components such as LSTMs that are capable of capturing temporal dependencies inherent in stress progression. In contrast, our results demonstrate that the temporal evolution of nitrogen stress is more accurately modeled using the sequential learning capacity of LSTMs, particularly when integrated with lightweight CNN backbones like MobileNetV2. Furthermore, the spatial feature representations extracted via MobileNetV2 showcased potential sensitivity to treatment-associated patterns during the evaluated acquisition period by capturing fine-grained morphological variations, which were not effectively addressed by the handcrafted features employed in Khanna et al.’s pipeline. This advantage is practically relevant during early phenological stages, where visible symptoms may be subtle, and precise morphological cues become essential for timely and accurate stress diagnosis.

\subsection{Performance Evaluation of Spatial Temporal Framework with Box-disjoint Approach}
The dataset was divided into mutually exclusive training and testing sets at the cultivation-box level, and the CNN–LSTM framework was evaluated according to Algorithm \ref{alg:lobo}. The model parameters were the same as those reported in Table \ref{tab:lstm_params}, except for four modifications: the image size was set to ($160 \times 160$) pixels, the batch size was reduced to 4, and the parameters "samples per window per class" and "minimum validation loss" were not applicable to this evaluation.

In each fold, one cultivation box from each nitrogen class was used for testing, while the remaining boxes were used for training, thereby ensuring a box-disjoint and approximately class-balanced evaluation. The cultivation-box stress conditions described in Section~2.1 are abbreviated as follows: SW = sufficient water, LW = low water, NW = no weeds, MW = medium weed pressure, and HW = high weed pressure.

\begin{table*}[ht]
\centering
\caption{Fold-wise box-disjoint test configuration and classification accuracy under different nitrogen, water, and weed-stress combinations.}
\label{tab:fold_wise_box_results}
\begin{tabular}{|c|c|c|c|c|}
\hline
\textbf{Fold} &
\textbf{Low-N Test Box} &
\textbf{Medium-N Test Box} &
\textbf{High-N Test Box} &
\textbf{Accuracy} \\
\hline
1 & 22: SW--NW & 13: LW--NW & 10: SW--HW & 0.8833 \\
\hline
2 & 23: SW--NW & 14: LW--NW & 11: SW--HW & 0.8583 \\
\hline
3 & 24: SW--NW & 15: LW--NW & 12: SW--HW & 0.8333 \\
\hline
4 & 25: SW--MW & 19: LW--HW & 16: SW--HW & 0.8500 \\
\hline
5 & 26: SW--MW & 20: LW--HW & 17: SW--HW & 0.5083 \\
\hline
6 & 27: SW--MW & 21: LW--HW & 18: SW--HW & 0.6750 \\
\hline
7 & 28: LW--NW & 4: SW--NW & 7: SW--MW & 0.5500 \\
\hline
8 & 29: LW--NW & 5: SW--NW & 8: SW--MW & 0.6083 \\
\hline
9 & 30: LW--NW & 6: SW--NW & 9: SW--MW & 0.7667 \\
\hline
\multicolumn{4}{|r|}{Mean Accuracy} & \textbf{0.7259} \\
\hline
\end{tabular}

\vspace{2mm}
\begin{minipage}{0.97\textwidth}
\footnotesize
\textit{Abbreviations:} N, nitrogen; SW, sufficient water; LW, low water; NW, no weeds; MW, medium weed pressure; HW, high weed pressure.
\end{minipage}
\end{table*}

The box-disjoint cross-validation achieved an overall mean accuracy of \(72.59\%\), with fold-wise accuracies ranging from \(50.83\%\) to \(88.33\%\). The model performed best in Folds~1--3, with an average accuracy of \(85.83\%\). In these folds, the three nitrogen classes were associated with relatively distinct combinations of water availability and weed pressure, which may have made their canopy and spectral patterns easier to distinguish.

Performance was comparatively lower in Folds~4--6, with a mean accuracy of \(67.78\%\). In these folds, all three nitrogen classes were exposed to weed competition, while the medium- and high-nitrogen treatments were both subjected to high weed pressure. These similar stress conditions may have produced overlapping plant responses, thereby making discrimination among the nitrogen classes more difficult. The considerable variation within this group, from \(50.83\%\) to \(85.00\%\), also indicates that the characteristics of individual cultivation boxes influenced model performance.

Folds~7--9 produced the lowest grouped mean accuracy of \(64.17\%\). In these folds, the low- and medium-nitrogen treatments were both maintained without weed competition but differed in water availability, whereas the high-nitrogen treatment was subjected to medium weed pressure under sufficient water conditions. The effects of low nitrogen combined with water limitation may have resembled those of high nitrogen combined with weed competition, resulting in partially overlapping plant-stress patterns.

The noticeable variation in accuracy across the nine folds demonstrates that both the combined-stress conditions and box-specific characteristics influenced model performance. Overall, the results indicate that the model was able to recognize nitrogen-associated stress patterns under several combined-stress conditions. However, the present experimental design does not allow nitrogen-specific effects to be completely separated from those caused by water limitation and weed competition.

\subsection{Performance Evaluation of Spatial Framework}

Before developing the final CNN--LSTM framework, we first evaluated lightweight CNN backbones to identify a suitable spatial feature extractor for nitrogen stress severity classification. Three lightweight architectures, namely \textit{MobileNetV2}, \textit{EfficientNetB0}, and \textit{NASNetMobile}, were explored. Among these, MobileNetV2 provided the best classification performance while retaining a compact architecture, and was therefore selected as the spatial feature extractor in the proposed CNN--LSTM framework. The implementation details and results of EfficientNetB0 and NASNetMobile are provided in Supplementary Material~S1.

To further quantify the contribution of spatio-temporal modeling vs spatial-only approach, MobileNetV2 framework was implemented as a baseline. In this setup, pretrained MobileNetV2 weights initialized from ImageNet were used, and the original classification head was removed so that the backbone could serve as a feature extractor. Custom dense layers were then added to adapt the model to the three-class nitrogen stress classification task. The model was trained for 250 epochs with a batch size of 64, providing stable gradient updates while maintaining computational efficiency. To address limited training data and simulate real-world variability, extensive data augmentation was applied through the \texttt{ImageDataGenerator} class with the following parameters: rescale = $1/255$, shear range = $0.2$, rotation range = $30^{\circ}$, width and height shift range = $0.2$, horizontal and vertical flips = \texttt{True}, and fill mode = \texttt{nearest}, thereby enhancing generalization and reducing overfitting.

For optimization, the Adam optimizer was used with an exponentially decaying learning rate schedule. The initial learning rate was set to $0.001$. The decay schedule followed an \texttt{ExponentialDecay} policy with $\text{decay\_steps} = \text{steps\_per\_epoch} \times 10$, $\text{decay\_rate} = 0.9$, and a staircase update, ensuring the learning rate decreased gradually as training progressed. This strategy stabilized convergence and avoided premature overfitting.

The network employed L2 regularization $(0.01)$ on the dense layers and dropout (rate = $0.5$) after each fully connected layer to further prevent overfitting. The training incorporated a \texttt{ModelCheckpoint} callback, saving the best-performing model weights per fold based on the lowest validation loss.

\begin{figure*}[hbtp]
    \centering

    \begin{subfigure}[b]{0.32\linewidth}
        \centering
        \includegraphics[width=\linewidth]{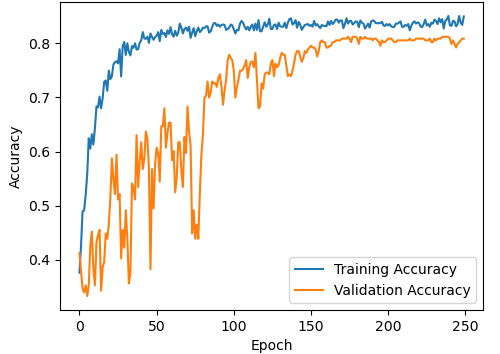}
        \caption{Fold 1}
        \label{fig:fold1_acc_s}
    \end{subfigure}
    \begin{subfigure}[b]{0.32\linewidth}
        \centering
        \includegraphics[width=\linewidth]{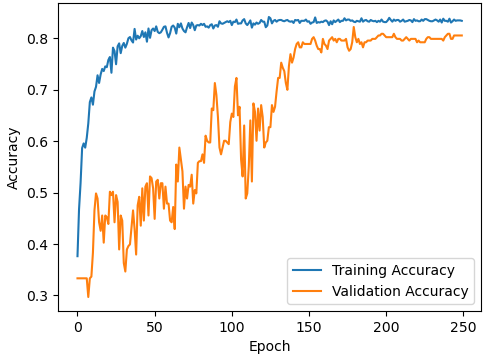}
        \caption{Fold 2}
        \label{fig:fold2_acc_s}
    \end{subfigure}
    \begin{subfigure}[b]{0.32\linewidth}
        \centering
        \includegraphics[width=\linewidth]{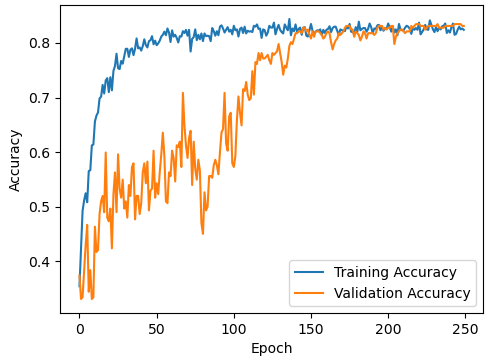}
        \caption{Fold 3}
        \label{fig:fold3_acc_s}
    \end{subfigure}

    \vspace{0.7em}

    \begin{subfigure}[b]{0.32\linewidth}
        \centering
        \includegraphics[width=\linewidth]{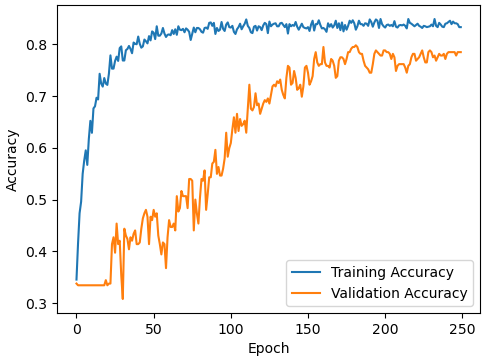}
        \caption{Fold 4}
        \label{fig:fold4_acc_s}
    \end{subfigure}
    \begin{subfigure}[b]{0.32\linewidth}
        \centering
        \includegraphics[width=\linewidth]{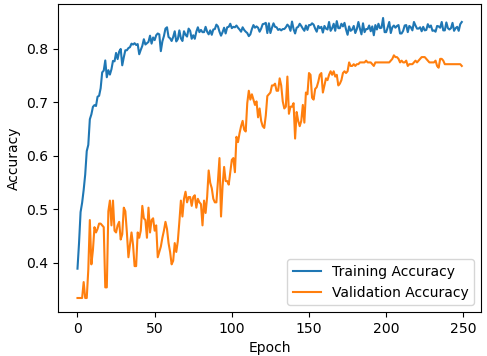}
        \caption{Fold 5}
        \label{fig:fold5_acc_s}
    \end{subfigure}

    \caption{Accuracy curves of Spatial Framework for 5-fold cross-validation (a–e correspond to Fold 1–5).}
    \label{fig:mobilenetv2_acc}
\end{figure*}

\begin{figure*}[hbtp]
    \centering

    \begin{subfigure}[b]{0.32\linewidth}
        \centering
        \includegraphics[width=\linewidth]{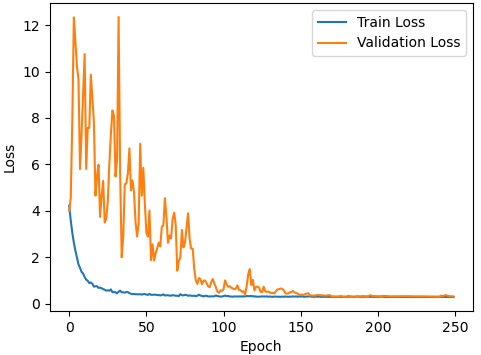}
        \caption{Fold 1}
        \label{fig:fold1_loss_s}
    \end{subfigure}
    \begin{subfigure}[b]{0.32\linewidth}
        \centering
        \includegraphics[width=\linewidth]{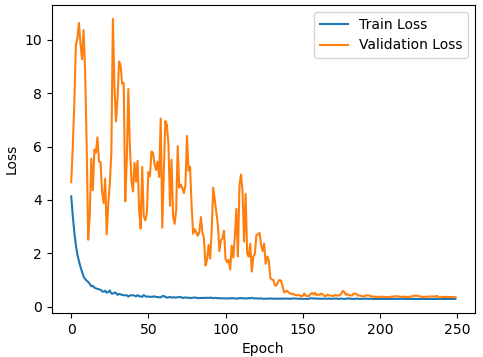}
        \caption{Fold 2}
        \label{fig:fold2_loss_s}
    \end{subfigure}
    \begin{subfigure}[b]{0.32\linewidth}
        \centering
        \includegraphics[width=\linewidth]{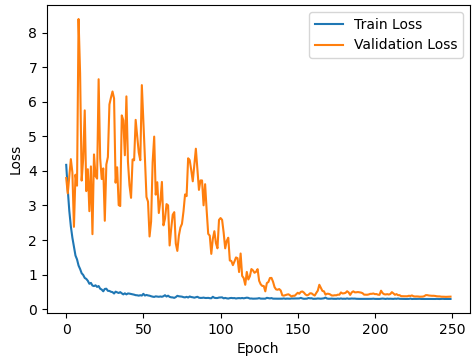}
        \caption{Fold 3}
        \label{fig:fold3_loss_s}
    \end{subfigure}

    \vspace{0.7em}

    \begin{subfigure}[b]{0.32\linewidth}
        \centering
        \includegraphics[width=\linewidth]{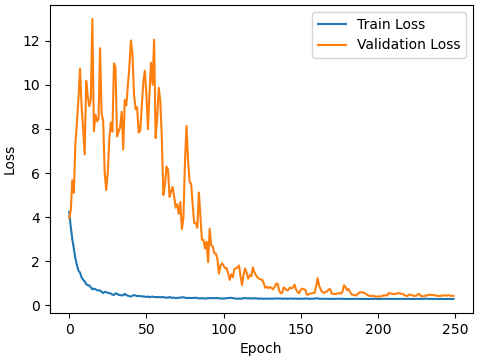}
        \caption{Fold 4}
        \label{fig:fold4_loss_s}
    \end{subfigure}
    \begin{subfigure}[b]{0.32\linewidth}
        \centering
        \includegraphics[width=\linewidth]{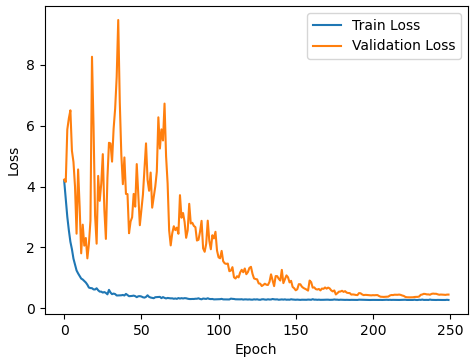}
        \caption{Fold 5}
        \label{fig:fold5_loss_s}
    \end{subfigure}

    \caption{Loss curves of Spatial Framework for 5-fold cross-validation (a–e correspond to Fold 1–5).}
    \label{fig:mobilenetv2_loss}
\end{figure*}

The model is trained and evaluated under a 5-fold stratified cross-validation strategy. Fig. \ref{frame_spatial}, \ref{fig:mobilenetv2_acc}, and \ref{fig:mobilenetv2_loss} illustrate the architecture, accuracy, and loss curves respectively. Table \ref{tab:cv_results} summarizes fold-wise training, validation, and test results, while Table~\ref{tab:prf_folds} reports the precision, recall, and F1-scores across classes.

\begin{table*}[htbp]
\centering
\caption{Training and validation performance across 5 folds in the spatial framework.}
\label{tab:cv_results}
\begin{tabular}{c c c c c c}
\hline
\textbf{Fold} & \textbf{Train Loss} & \textbf{Train Acc.} &
\textbf{Val. Loss} & \textbf{Val. Acc.} & \textbf{Epochs} \\
\hline
1 & 0.2855 & 0.8412 & 0.2949 & 0.8119 & 238 \\
2 & 0.2943 & 0.8337 & 0.3470 & 0.8053 & 240 \\
3 & 0.3022 & 0.8165 & 0.3636 & 0.8344 & 245 \\
4 & 0.2830 & 0.8322 & 0.3968 & 0.7881 & 200 \\
5 & 0.2747 & 0.8331 & 0.3605 & 0.7848 & 224 \\
\hline
\textbf{Mean} & 0.2879 & 0.8313 & 0.3526 & 0.8049 & -- \\
\textbf{Std}  & 0.0098 & 0.0081 & 0.0370 & 0.0186 & -- \\
\hline
\end{tabular}
\end{table*}

As illustrated in Fig. \ref{fig:mobilenetv2_acc}, across all folds, the training accuracy rapidly converged to approximately 0.83, while validation accuracy improved gradually and stabilized in the range of 0.78--0.83. Fold~3 demonstrated the strongest alignment between training and validation curves, whereas Folds~4 and~5 displayed a slightly larger gap, suggesting mild overfitting. The corresponding loss curves (Fig. \ref{fig:mobilenetv2_loss}) revealed sharp decreases in training loss, with validation loss exhibiting high variance in the early epochs but stabilizing after 150 epochs. These observations confirm that data augmentation, dropout, and L2 regularization were effective in mitigating overfitting, while the exponentially decaying learning rate ensured stable convergence.

\begin{table*}[htbp]
\centering
\caption{Precision, Recall, and F1-score across 5 folds in Spatial Framework}
\label{tab:prf_folds}
\begin{tabular}{lcccccc}
\hline
\textbf{Class} & \textbf{Metric} & \textbf{Fold 1} & \textbf{Fold 2} & \textbf{Fold 3} & \textbf{Fold 4} & \textbf{Fold 5} \\
\hline
\multirow{3}{*}{High} 
 & Precision & 0.64 & 0.99 & 0.97 & 0.99 & 0.99 \\
 & Recall    & 1.00 & 0.72 & 0.75 & 0.73 & 0.69 \\
 & F1-score  & 0.78 & 0.83 & 0.85 & 0.84 & 0.81 \\
\hline
\multirow{3}{*}{Low} 
 & Precision & 1.00 & 0.64 & 0.96 & 0.97 & 0.61 \\
 & Recall    & 0.72 & 0.96 & 0.76 & 0.65 & 0.99 \\
 & F1-score  & 0.84 & 0.77 & 0.85 & 0.78 & 0.76 \\
\hline
\multirow{3}{*}{Medium} 
 & Precision & 1.00 & 0.96 & 0.69 & 0.62 & 0.99 \\
 & Recall    & 0.71 & 0.73 & 0.99 & 0.98 & 0.67 \\
 & F1-score  & 0.83 & 0.83 & 0.81 & 0.76 & 0.80 \\
\hline
\end{tabular}
\end{table*}

Table~\ref{tab:cv_results} shows that the model achieved an average training accuracy of 83.13\% and a validation accuracy of 80.49\% across folds, with low standard deviation (1.86\%). The highest validation accuracy was observed in Fold~3 (83.44\%), while the lowest occurred in Fold~5 (78.48\%). Training and validation losses were consistent across folds, with only minor fluctuations.

As shown in Table~\ref{tab:prf_folds}, class-wise evaluation revealed balanced predictive capacity, with F1-scores ranging from 0.76 to 0.85 across all classes and folds. For the \textit{High} class, precision was strong in most folds ($\geq 0.97$) but recall was comparatively lower (0.69--0.75), except in Fold~1 where recall reached 1.00. Conversely, the \textit{Low} class showed complementary trends, with high recall in Folds~2 and~5 ($\geq 0.95$) but reduced precision (0.61--0.64). The \textit{Medium} class was the most variable, with precision fluctuating between 0.62 and 1.00, though recall generally remained high (0.71--0.99). These results suggest that inter-class boundaries are occasionally ambiguous, leading to trade-offs in precision and recall.

Overall, the spatial framework achieved a mean validation/evaluation accuracy of 80.49\%, with relatively consistent performance across the five folds and per-class F1-scores ranging from 0.76 to 0.85. Although these results demonstrate the usefulness of spatial canopy features for classifying nitrogen-treatment levels, the observed variation in class-specific precision and recall indicates that static structural and colour-related information may not always be sufficient to distinguish treatment classes under overlapping water and weed-stress conditions. The CNN--LSTM framework achieved a higher mean validation/evaluation accuracy of 95.37\% under the forward-chaining configuration, corresponding to an absolute difference of 14.88 percentage points relative to the spatial baseline. However, this difference should be interpreted cautiously because the two frameworks were evaluated using different sequence-construction procedures, cross-validation strategies, training configurations, and experimental units.

\subsection{Comparison with Machine Learning Methods}
Notably, this performance surpasses that of traditional models, as reported by Khanna et al.\cite{khanna_spatio_2019}. The comparative analysis of performances is presented in Table \ref{tab:comp_ML_CNN_LSTM}. To contextualize our findings, we compared the performance of proposed framework against that of conventional machine learning models and the spatial-only CNN. Traditional classifiers such as Decision Trees, KNN, and Bagged Trees achieved test accuracies below 70\%, while SVM achieved a higher accuracy of 80.95\%, comparable to the spatial-only CNN.
In contrast, the spatio-temporal CNN-LSTM framework markedly outperformed all baselines, achieving 95.37\% test accuracy. This performance gain demonstrates that spatio-temporal modeling provides a substantial advantage in resolving confounding stress symptoms and effectively predicting nitrogen severity classes. The box-disjoint cross-validation yielded the mean training accuracy of \(77.23\%\) and an overall mean accuracy of \(72.59\%\), 
with fold-wise accuracies ranging from \(50.83\%\) to \(88.33\%\).
Reduced performance in some folds suggests that 
overlapping plant responses caused by drought and weed competition made the 
nitrogen classes more difficult to separate.

\begin{table*}[htbp]
\centering
\setlength{\tabcolsep}{4pt}
\renewcommand{\arraystretch}{1.15}
\caption{Comparison of classification accuracy between conventional machine-learning classifiers reported by Khanna et al.~\cite{khanna_spatio_2019} and the proposed spatial and spatio-temporal frameworks under the three evaluation protocols used in this study.}
\label{tab:comp_ML_CNN_LSTM}
\begin{tabularx}{\textwidth}{
    @{}
    >{\raggedright\arraybackslash}X
    >{\centering\arraybackslash}p{2.8cm}
    >{\centering\arraybackslash}p{2.6cm}
    >{\centering\arraybackslash}p{1.8cm}
    @{}
}
\hline
\textbf{Method}
&
\makecell{\textbf{Mean Accuracy}\\
          \textbf{Train (\%)}}
&
\makecell{\textbf{Mean Accuracy}\\
          \textbf{Validation/Test (\%)}}
&
\textbf{Reference}
\\
\hline

Decision Trees
& 63.66 & 47.62
& \multirow{8}{*}{\cite{khanna_spatio_2019}}
\\

LDA
& 68.47 & 78.57 & \\

SVM
& 75.68 & 80.95 & \\

KNN
& 62.16 & 55.95 & \\

Bagged Trees
& 67.57 & 63.10 & \\

Subspace Discriminant
& 70.57 & 75.00 & \\

Subspace KNN
& 60.66 & 66.67 & \\

RUSBoosted Trees
& 69.37 & 63.10 & \\
\hline

Proposed Spatial Framework
& 83.13 & 80.49
& \multirow{3}{*}{Proposed}
\\

Proposed Spatio-Temporal Framework
(Forward Chaining)
& 98.62 & 95.37 & \\

Proposed Spatio-Temporal Framework
(Box-Disjoint)
& 77.23 & 72.59 & \\
\hline

\end{tabularx}
\end{table*}

Khanna et al.\cite{khanna_spatio_2019} rely on explicitly derived plant trait indicators—canopy cover, height, spectral indices, reflectance statistics, and 3D point-cloud features as inputs to classical ML classifiers (SVM, RF, KNN, etc.). In contrast, our DL frameworks learn discriminative spatial features directly from raw multi-modal images, eliminating the need for domain-specific handcrafted traits and enabling more scalable deployment across crops and environments. Thus, beyond the expected accuracy gap, Table \ref{tab:comp_ML_CNN_LSTM} highlights that traditional ML models are fundamentally constrained by handcrafted feature dependence, weak temporal modelling, and reduced sensitivity to overlapping stress phenotypes. Our deep learning–based spatio-temporal frameworks address these limitations by learning hierarchical representations directly from raw data and explicitly modelling stress dynamics over time, leading to substantially improved precision and accuracy under combined stress conditions.

\subsection{Practical Implications}
From a practical standpoint, the findings provide useful guidance on when the model may support nitrogen-management decisions and how it could be deployed in the field. The forward-chaining results showed consistently high classification performance across all six validation folds, indicating that nitrogen-treatment classes were already distinguishable at the first validation anchor. However, because earlier acquisition dates were included only as temporal context and were not independently evaluated as validation anchors, the earliest possible time for reliable nitrogen-management intervention cannot be determined from the present analysis. The framework also works directly with raw multimodal canopy images, including RGB, infrared, and multi-spectral data, without relying on derived vegetation indices or three-dimensional point-cloud features. This simplifies the processing workflow and may improve compatibility with a wider range of sensing platforms. In addition, the lightweight MobileNetV2 architecture has potential for edge deployment on drones or ground-based rovers, which could enable near-real-time monitoring across larger cultivation areas. However, further field-based evaluation is needed to confirm its consistency, computational efficiency, and scalability across different environments, crop growth stages, sensors, and hardware configurations.

\section{Conclusion}

The findings demonstrate that nitrogen-stress severity under combined drought and weed pressure is expressed through both instantaneous spatial patterns and their progression over time. The spatial-only \textit{MobileNetV2} framework achieved a mean accuracy of approximately 80.49\%, indicating that raw RGB, infrared, and multi-spectral images contain discriminative canopy-level information sufficient to distinguish nitrogen treatments even when their visible responses are modified by water limitation and weed competition. Importantly, this framework learns relevant spatial representations directly from the images, eliminating the need for domain-specific handcrafted traits---such as canopy cover, plant height, spectral indices, reflectance statistics, and 3D point-cloud features---as inputs to classical machine-learning classifiers. The significance of the spatial-only approach therefore extends beyond its role as a baseline: it demonstrates that useful nitrogen-stress information can be extracted directly from multimodal imagery, potentially simplifying model adaptation and supporting more scalable deployment across crops, sensors, and environments.

The higher forward-chaining accuracy of the CNN--LSTM framework (95.37\%) supports hypothesis H$_1$ and shows that temporal sequences provide information beyond static canopy appearance. Mechanistically, the CNN captures spatial manifestations of stress, including differences in canopy structure, colour, and spectral response, while the LSTM integrates how these manifestations develop across successive acquisition dates. Temporal progression therefore helps distinguish nitrogen-associated responses from transient or visually similar effects induced by drought and weed competition. The stable performance across successive validation anchors further indicates that these stress-progression signatures remain detectable across the evaluated growth-stage intervals.

The box-disjoint accuracy of 72.59\%, which was lower and more variable than the forward-chaining result, supports hypothesis H$_2$. This outcome indicates that temporal nitrogen-stress signatures generalize less consistently when the model encounters unseen cultivation boxes with different combinations of nitrogen supply, water availability, weed pressure, and box-specific characteristics. Thus, the three analyses provide complementary evidence: spatial learning establishes the diagnostic value of single-date raw imagery, forward chaining evaluates the detectability of treatment-associated patterns at successive acquisition-date anchors, and box-disjoint evaluation identifies the principal barrier to deployment---generalization under overlapping and previously unseen treatment conditions. Collectively, these findings support lightweight, trait-free spatio-temporal learning as a scalable basis for precision nitrogen monitoring, while emphasizing the need for broader field, crop, seasonal, and environmental validation before operational application.

\section*{Declaration of Competing Interest}
The authors declare that they have no known competing financial interests or personal relationships that could have appeared to influence the work reported in this paper.

\section*{CRediT authorship contribution statement}
\textbf{Aswini Kumar Patra:} Conceptualization, Methodology, Resources, Software, Writing-Original Draft, Investigation, Formal Analysis, Visualization \& Validation; \textbf{Anshu Rastogi:} Methodology, Review \& Editing; \textbf{Lingaraj Sahoo:} Review, Editing \& Supervision.

\sloppy
\section*{Data Availability}
The data used in this study are derived from a publicly available dataset, which can be accessed at \url{https://projects.asl.ethz.ch/datasets/plant-stress-phenotyping-2018/}. Additionally, the generated data supporting the findings of this study are available from the corresponding author upon reasonable request.
\section*{Funding}
The authors declare that no funding was received for this research.


%
%
%
\bibliographystyle{unsrt}
\bibliography{ref}

\end{document}